\documentclass[11pt]{article}

\usepackage{fullpage}
\usepackage[ruled, linesnumbered, vlined, commentsnumbered]{algorithm2e}
\usepackage{graphicx,subfig} 
\usepackage{amsfonts,amssymb,amsmath,amsthm,amsopn,mathrsfs,mathtools,wasysym}	
\usepackage{hhline,booktabs,colortbl,multirow,tabularx,diagbox,threeparttable} 
\usepackage[listings,skins,breakable]{tcolorbox}

\usepackage{enumerate}
\usepackage{authblk}
\usepackage{footnote}
\usepackage{hyperref}
\usepackage{prettyref}
\usepackage{cite}
\usepackage{setspace}
\usepackage{color}
\usepackage{xcolor}  
\usepackage{scrpage2}
\usepackage{geometry}

\usepackage{tikz}
\usepackage{pgfplots}
\usetikzlibrary{positioning,shapes,shadows,arrows,calc}
\tikzstyle{component}=[rectangle, draw=black, rounded corners, fill=blue!40, drop shadow, text centered, anchor=north, text=white, minimum height=1cm]
\tikzstyle{arrow}=[->, thick]

\pgfplotsset{compat=1.12}
\usetikzlibrary{intersections}
\usetikzlibrary{pgfplots.statistics}
\usepgfplotslibrary{fillbetween}

\definecolor{myblue}{RGB}{34,31,217}
\definecolor{mycyan}{gray}{.7}
\definecolor{Gray}{gray}{0.9}

\newtheorem{remark}{Remark}

\newcommand{\pref}{\prettyref}

\newrefformat{fig}{Fig.~\ref{#1}}
\newrefformat{tab}{Table~\ref{#1}}
\newrefformat{sec}{Section~\ref{#1}}
\newrefformat{alg}{Algorithm~\ref{#1}}
\newrefformat{property}{Property~\ref{#1}}
\newrefformat{theorem}{Theorem~\ref{#1}}
\newrefformat{definition}{Definition~\ref{#1}}
\newrefformat{corollary}{Corollary~\ref{#1}}
\newrefformat{lemma}{Lemma~\ref{#1}}
\newrefformat{conj}{Conjecture~\ref{#1}}
\newrefformat{def}{Definition~\ref{#1}}
\newrefformat{eq}{equation~(\ref{#1})}
\newrefformat{app}{Appendix~\ref{#1}}

\usepackage{lscape}

\begin{document}

\title{\vspace{-1ex}\LARGE\textbf{Interactive Evolutionary Multi-Objective Optimization via Learning-to-Rank}~\footnote{This manuscript is submitted for potential publication. Reviewers can use this version in peer review.}}

\author[1]{\normalsize Ke Li}
\author[2]{\normalsize Guiyu Lai}
\author[2]{\normalsize Xin Yao}
\affil[1]{\normalsize Department of Computer Science, University of Exeter, EX4 4QF, Exeter, UK}
\affil[2]{\normalsize College of Computer Science \& Engineering, University of Electronic Science and Technology of China, Chengdu 611731, China}
\affil[3]{\normalsize Department of Computer Science and Engineering, Southern University of Science and Technology, Shenzhen 518055, China}
\affil[$\ast$]{\normalsize Email: \texttt{k.li@exeter.ac.uk}}

\date{}
\maketitle

\vspace{-3ex}
{\normalsize\textbf{Abstract: } }In practical multi-criterion decision-making, it is cumbersome if a decision maker (DM) is asked to choose among a set of trade-off alternatives covering the whole Pareto-optimal front. This is a paradox in conventional evolutionary multi-objective optimization (EMO) that always aim to achieve a well balance between convergence and diversity. In essence, the ultimate goal of multi-objective optimization is to help a decision maker (DM) identify solution(s) of interest (SOI) achieving satisfactory trade-offs among multiple conflicting criteria. Bearing this in mind, this paper develops a framework for designing preference-based EMO algorithms to find SOI in an interactive manner. Its core idea is to involve human in the loop of EMO. After every several iterations, the DM is invited to elicit her feedback with regard to a couple of incumbent candidates. By collecting such information, her preference is progressively learned by a learning-to-rank neural network and then applied to guide the baseline EMO algorithm. Note that this framework is so general that any existing EMO algorithm can be applied in a plug-in manner. Experiments on $48$ benchmark test problems with up to $10$ objectives fully demonstrate the effectiveness of our proposed algorithms for finding SOI.



\section{Introduction}
\label{sec:introduction}


Partially due to the population-based property, evolutionary algorithms (EAs) have been widely recognized to be effective for multi-objective optimization problems (MOPs). Over the past three decades and beyond, many efforts have been dedicated to developing evolutionary multi-objective optimization (EMO) algorithms to obtain a set of trade-off solutions that approximate a Pareto-optimal front (PF) with a decent diversity. Existing EMO algorithms can be classified into three categorizes, i.e., Pareto-, indicator-, and decomposition-based EMO approaches, where fast and elitist multi-objective genetic algorithm (NSGA-II)~\cite{DebAPM02,LiKCLZS12}, indicator-based EA (IBEA)~\cite{ZitzlerK04} and multi-objective EA based on decomposition (MOEA/D)~\cite{ZhangL07,PruvostDLL020} are representative algorithms respectively. Approximating the entire PF can be a double-edged sword when it is handed over to the decision maker (DM) at the \textit{posteriori} multi-criterion decision-making (MCDM) stage. Due to the negligence of DM's preference in the loop, it is not guaranteed to identify the solution(s) of interest (SOI) most relevant to the DM's aspiration. This is further aggravated when the number of objectives becomes large given that the PF approximation is too sparse to cover the SOI, letting alone the cognitive barrier for understanding and interpreting the high-dimensional data.

Comparing to the \textit{posteriori} decision-making, a range of empirical studies reported in our recent paper~\cite{LiLDMY20} have shown the benefits of incorporating the DM's preference information into the EMO process for locating the SOI. According to the preference elicitation manner, i.e., when to ask the DM to elicit/input her preference information, the existing preference-based EMO algorithms can be categorized into \textit{a priori} EMO and \textit{interactive} EMO. In view of the black box nature of real-world problems, of which the DM has little knowledge, it is controversial to elicit a reasonable preference information beforehand. Even worse, a disruptive preference information can lead to a failure of the underlying algorithm.

In contrast, the \textit{interactive} EMO (e.g.,~\cite{MiettinenM00,Li19,LiCSY19,DebK07CEC,BrankeGSZ15,TomczykK20,LaiL021,XuLA22}) provides a better opportunity for the DM to progressively understand the underlying black box system thus to gradually amend her preference information. The DM is involved in the overall \textit{optimization-cum-decision-making} loop as being periodically requested to input her preference information w.r.t. the \textit{selected} solutions provided by the underlying EMO algorithm. Both direct (e.g., weights or an aspiration level vector~\cite{DebK07CEC}) and indirect information (e.g., score~\cite{LiCSY19} or holistic pairwise comparison~\cite{DebSKW10,BrankeGSZ15,TomczykK20}) can be used to represent the DM's preference. As discussed in~\cite{LiLDMY20}, eliciting direct preference information is far from trivial when encountering a black box system. It is likely to be error prone and cognitively demanding when the number of objectives becomes large. In contrast, the indirect information have become more appealing and prevalent in the interactive EMO literature. Based on the inputs/feedback collected from the DM, a preference model is progressively learned (e.g., radial basis function networks~\cite{LiCSY19}, nonlinear programming~\cite{DebSKW10} and ordinal regression~\cite{BrankeGSZ15}), and it is used to guide the population towards the SOI.

Note that interactive multi-objective optimization approaches have been a longstanding area in the MCDM community~\cite{nonlinear} decades before the emergence of the EMO. In recent years, we have witnessed a growing trend of seeking synergies between EMO and MCDM. For example, our recent work~\cite{LiCSY19} developed a framework for designing interactive EMO approaches via preference learning and it showed encouraging results on problems with up to $10$ objectives. Unfortunately, this work is merely valid for decomposition-based EMO approaches and it is yet applicable for the Pareto- and indicator-based EMO approaches. In fact, most, if not all, existing interactive EMO approaches are designed to be algorithm-specific (e.g.,~\cite{DebSKW10} and~\cite{BrankeGSZ15} are designed for NSGA-II, \cite{ChughSHM15} is designed for IBEA, \cite{LiCSY19} and~\cite{TomczykK20} are designed for MOEA/D). The prevalent preference learning, from indirect information, in interactive EMO is mainly derived from mathematical programming and operations research approaches while machine learning approaches such as learning-to-rank (LTR)~\cite{Liu09} has rarely been considered, except~\cite{BattitiP10}. Note that LTR has been recognized as an effective tool to learn user preference in information retrieval and recommendation systems~\cite{GemmisILMNS10}.

Built upon our previous work~\cite{LiCSY19}, this paper develops a general framework to design interactive EMO algorithms that progressively learn the DM's preference information from her feedback and adapt the learned preference to guide the population towards the SOI. As in~\cite{LiCSY19}, this framework consists of three modules, i.e., \texttt{consultation}, \texttt{preference elicitation} and \texttt{optimization}.
\begin{itemize}
    \item As the interface by which the DM interacts with the EMO algorithm, the \texttt{consultation} module mainly aims to collect the DM's preference information from her feedback and to build a preference model. In particular, this paper considers an indirect preference information in the form of holistic pairwise comparisons of solutions. The preference model is built upon the collected comparison results, constituting the training data, by using a LTR neural network.
    \item In view of the unique environmental selection mechanism of the underlying EMO algorithm, the preference model is usually not directly applicable. The \texttt{preference elicitation} module plays as the catalyst that translates the preference information learned in the \texttt{consultation} module into the form that can be used in the underlying EMO algorithm.
    \item The \texttt{optimization} module can be any EMO algorithm in the literature. For proof-of-concept purposes, this paper chooses three iconic algorithms from each of Pareto-, indicator- and decomposition-based EMO approaches (i.e., NSGA-II, IBEA and MOEA/D). The \texttt{optimization} module leverages the preference information learned from the \texttt{preference elicitation} module to search for the SOI. In the meanwhile, it periodically provides the \texttt{consultation} module with a set of selected candidates for preference learning.
    \item To validate the effectiveness of our proposed framework, we instantiate three interactive EMO algorithms, denoted as \texttt{I-NSGA-II/LTR}, \texttt{I-R2-IBEA/LTR} and \texttt{I-MOEA/D/LTR}. We compare their performance against four state-of-the-art (SOTA) interactive EMO algorithms, \texttt{BC-EMO}~\cite{BattitiP10}, \texttt{NEMO-0}~\cite{BrankeGSZ15}, \texttt{I-MOEA/D-PLVF}~\cite{LiCSY19}, and \texttt{IEMO/D}~\cite{TomczykK20}, on $48$ benchmark problem instances with a range of different characteristics.
\end{itemize}

For the remaining parts of this paper, \pref{sec:preliminaries} provides some necessary background knowledge including a pragmatic overview of the related works of preference-based EMO in an interactive manner. \pref{sec:method} delineates the implementation detail of the proposed interactive EMO framework. \pref{sec:settings} introduces the experimental settings while the comparison results w.r.t. SOTA peer algorithms are presented and discussed in~\pref{sec:experiments}. At the end, \pref{sec:conclusion} concludes this paper and sheds some lights on future directions.


\section{Preliminaries}
\label{sec:preliminaries}

This section starts with some basic definitions related to this paper, followed by a pragmatic overview on some selected developments of preference-based EMO in an interactive manner. Interested readers are referred to some recent survey papers~\cite{WangOJ17,XinCCIHL18} and~\cite{LiLDMY20} for a more comprehensive survey.

\subsection{Basic Definitions}
\label{sec:definitions}

The MOP considered in this paper is formulated as:
\begin{equation}
\begin{array}{l}
\mathrm{minimize} \quad \mathbf{F}(\mathbf{x})=(f_1(\mathbf{x}),\cdots,f_m(\mathbf{x}))^{T}\\
\mathrm{subject\ to} \quad \mathbf{x}\in\Omega
\end{array},
\label{eq:MOP}
\end{equation}
where $\mathbf{x}=(x_1,\cdots,x_n)^T$ is a $n$-dimensional decision vector and $\mathbf{F}(\mathbf{x})$ is an $m$-dimensional objective vector. $\Omega$ is the feasible set in the decision space $\mathbb{R}^n$ and $\mathbf{F}:\Omega\rightarrow\mathbb{R}^m$ is the corresponding attainable set in the objective space $\mathbb{R}^m$. Without considering the DM's preference information, given two solutions $\mathbf{x}^1,\mathbf{x}^2\in\Omega$, $\mathbf{x}^1$ is said to dominate $\mathbf{x}^2$ if and only if $f_i(\mathbf{x}^1)\leq f_i(\mathbf{x}^2)$ for all $i\in\{1,\cdots,m\}$ and $\mathbf{F}(\mathbf{x}^1)\neq\mathbf{F}(\mathbf{x}^2)$. A solution $\mathbf{x}\in\Omega$ is said to be Pareto-optimal if and only if there is no solution $\mathbf{x}^\prime\in\Omega$ that dominates it. The set of all Pareto-optimal solutions is called the Pareto-optimal set (PS) and their corresponding objective vectors constitute the PF. Accordingly, the ideal point is defined as $\mathbf{z}^{\ast}=(z^{\ast}_1,\cdots,z^{\ast}_m)^T$, where $z^{\ast}_i=\min\limits_{\mathbf{x}\in\Omega}f_i(\mathbf{x})$, and the nadir point is defined as $\mathbf{z}^{\texttt{nad}}=(z^{\texttt{nad}}_1,\cdots,z^{\texttt{nad}}_m)^T$, where $z^{\texttt{nad}}_i=\max\limits_{\mathbf{x}\in PS}f_i(\mathbf{x})$, $\forall i\in\{1,\cdots,m\}$.

\subsection{Related Works on Interactive EMO}
\label{sec:related}

\begin{table*}[t!]
\centering
\caption{Summary of the key features of selected interactive EMO algorithms.}
\resizebox{\textwidth}{!}{
    \begin{tabular}{c|c|c|c|c|c}
    \hline
        \multirow{2}{*}{\textsc{Algorithm}} & \multirow{2}{*}{\textsc{Optimization Module}} & \multicolumn{2}{c|}{\textsc{Consultation Module}} & \multirow{2}{*}{\textsc{Elicitation Module}} & \multirow{2}{*}{$m$} \\\cline{3-4}
        &       & \textsc{Preference Information} & \textsc{Preference Model} &       &  \\\hline\hline
        IEM~\cite{PhelpsK03}   & Genetic algorithm & Pairwise comparisons & Additive linear value function & Ranking based on utility function & $4$ \\\hline
        PI-NSGA-II-VF~\cite{DebSKW10} & NSGA-II & Pairwise comparisons & Polynomial value function & Modified dominance principle & $5$ \\\hline
        BC-EMOA~\cite{BattitiP10} & NSGA-II & Pairwise comparisons & Support vector machine & Ranking based on utility function & $10$ \\\hline
        NEMO-0, NEMO-I, NEMO-II~\cite{BrankeGSZ15} & NSGA-II & Pairwise comparisons & Ordinal regression & Ranking based on utility function & $5$ \\\hline
        IEMO/D~\cite{TomczykK20} & MOEA/D & Pairwise comparisons & $L_\alpha$-norm & Biased weight vectors & $15$ \\\hline
        PC~\cite{ParmeeC02}    & Genetic algorithm & Objective comparisons & Fuzzy logic & Ranking based on utility function & $2$ \\\hline
        RWA~\cite{JinS02}   & Genetic algorithm & Objective comparisons & Fuzzy logic & Ranking based on utility function & $2$ \\\hline
        FLMOEA~\cite{ShenGCH10} & NSGA-II & Objective importance & Fuzzy logic & Ranking based on utility function & $7$ \\\hline
        WWW-NIMBUS~\cite{MiettinenM00} & Nonlinear programming & Objective importance & Achievement scalarizing function & Ranking based on utility function & \diagbox{}{} \\\hline
        NAUTILUS~\cite{MiettinenERL10}  & Nonlinear programming & Objective importance & Achievement scalarizing function & Ranking based on utility function & $4$ \\\hline
        PIE~\cite{SindhyaRM11}   & Evolutionary algorithm & Objective importance & Achievement scalarizing function & Ranking based on utility function & $5$ \\\hline
        GZGZY~\cite{GuoZGZY20} & NSGA-II & Semantic-based relative importance & Preference regions & Preference regions & $3$ \\\hline
        GRIST~\cite{Yang99,YangL02,LuqueYW09} & Nonlinear programming & Trade-off relation & Utility function & Utility function & $3$ \\\hline
        T-IMO-EA~\cite{ChenXC17} & Evolutionary algorithm & Trade-off relation & Utility function & Normal vector of the tangent hyperplane of the PF & $5$ \\\hline
        FGKKMW~\cite{FowlerGKKMW10} & Genetic algorithm & Best and worst solution(s) & Polyhedral cones & Modified dominance principle & $4$ \\\hline
        PI-NSGA-II-PC~\cite{SinhaKWD10} & NSGA-II & Most preferred solution(s) & Polyhedral cones & Modified dominance principle & $5$ \\\hline
        IEA~\cite{GongSJ13}   & Evolutionary algorithm & Worst and best solutions & Polyhedron & Ranking based on preference polyhedron & $5$ \\\hline
        NEMO-II-Ch~\cite{BrankeCGSZ16} & NSGA-II & pairwise comparisons & Choquet integral & Ranking based on utility function & $5$ \\\hline
        iTDEA~\cite{KarahanK10} & TDEA  & Best solution & Favorable weights & Preferred weight region & $5$ \\\hline
        iMOEA/D~\cite{GongLZJZ11} & MOEA/D & Best solution & Neighborhood & Biased weight vectors & $3$ \\\hline
        I-MOEA/D-PLVF~\cite{LiCSY19} & MOEA/D & Performance score & Radial basis function network & Biased weight vectors & $10$ \\\hline
        \end{tabular}
}
\label{tab:literature}
\end{table*}

The initial attempt to incorporate the DM's preference into EMO can be traced back to the early $90$s when Fonseca and Fleming~\cite{FonsecaF98} suggested to model the DM’s preference as a goal that indicates desirable levels of performance at each objective dimension. The early works on preference-based EMO mainly use \textit{a priori} preference information where the DM only \lq interact\rq\ with the algorithm once at the outset of evolution. Almost all \textit{a priori} preference-based EMO approaches can be applied in an interactive manner. For example, the DM can periodically adjust the reference point to progressively guide the population towards the SOI. However, partially due to the use of direct preference information, such as reference point (also known as aspiration level vector)~\cite{DebSBC06,ThieleMKL09,SaidBG10,LiCMY18}, weights~\cite{BrankeKS01,Deb03,ZitzlerBT06} and desirability function~\cite{WagnerT10}, these approaches are cognitively demanding and highly likely to be error prone when the DM is intensively involved in the optimization loop.

In contrast, the interactive EMO, which bridges the gap between two sibling communities, i.e., EMO and MCDM, endeavor to keep DMs in the optimization loop thus to enable a collaborative human-computer optimization paradigm. The key features of some important developments of interactive EMO are summarized in~\pref{tab:literature}. The following paragraphs elaborate upon them according to the types of preference information, all of which are in an indirect format.

\subsubsection{Holistic pairwise comparisons}

During the consultation stage, the DM is asked to input her preference over a pair of candidates $\langle\mathbf{x}^a,\mathbf{x}^b\rangle$ at a time, such as $\mathbf{x}^a$ is better, worse or indifferent over $\mathbf{x}^b$. As a pioneer along this line, Phelps and K\"oksalan~\cite{PhelpsK03} proposed an interactive evolutionary meta-heuristic algorithm that translates the pairwise comparison results into an utility function as a weighted sum. In particular, the weights therein are estimated by solving a linear programming problem. Likewise, Deb et al. developed an interactive NSGA-II that progressively learns an approximated value function by asking the DM to compare a set of solutions in a pairwise manner~\cite{DebSKW10}. In~\cite{BattitiP10}, Battiti and Passerini proposed to use a high order polynomial as the utility function whose parameters are estimated by a support vector machine. The approximated utility function is used to modify the Pareto dominance in the environmental selection of NSGA-II. In~\cite{BrankeGSZ15}, Branke et al. proposed to use a robust ordinal regression to learn a representative additive monotonic value function that is used to replace the crowding distance in NSGA-II. Later, the same authors proposed an improved version of NEMO-II that applies Choquet integral as the DM's preference model~\cite{BrankeCGSZ16}. Recently, Tomczyk and Kadzi\'nski~\cite{TomczykK20} proposed an interactive EMO based on MOEA/D. It employs the $L_\alpha$-norm as the preference model and uses a Monte Carlo simulation with a rejection sampling to generate a set of weight vectors compatible with the learned preference information.

\subsubsection{Objective-level comparisons}

This type of preference information mainly exploit the relationship and importance among different objectives contingent upon the DM's preference. The first attempt along this line is~\cite{ParmeeC02} where Parmee and Cvetkovi\'c developed a fuzzy preference relation that translates the pairwise comparisons among objectives into a weighted-dominance relation according to the relative importance of objectives. By comparing objectives, Jin and Sendhoff~\cite{JinS02} proposed to use a fuzzy logic to convert the DM's preference information into weight intervals. In~\cite{ShenGCH10}, Shen et al. proposed an interactive EMO algorithm that periodically asks the DM to specify the relative importance between pairs of objectives via linguistic terms. Thereafter, the collected preference information is used to construct a new fitness function derived from a fuzzy inference system. By asking the DM to classify objectives into up to five classes, Miettinen and M{\"{a}}kel{\"{a}} developed an interactive multi-objective optimization system dubbed WWW-NIMBUS to progressively search for the SOI~\cite{MiettinenM00}. Later, Miettinen et al.~\cite{MiettinenERL10} proposed the NAUTILUS method that starts the search process from the estimated nadir point and interactively improve all objectives. Note that the DM is able to control the interaction frequency and improvement rates at different objectives. In~\cite{SindhyaRM11}, Sindhya et al. proposed to use an EA as the search engine in the NAUTILUS. Recently, Guo et al.~\cite{GuoZGZY20} proposed an interactive EMO algorithm that uses a three-step process, called partitioning-updating-tracking, to search for the SOI. In particular, the quality of a solution is measured according to the satisfaction of the semantic-based relative importance of different objectives.

\subsubsection{Trade-off specification}

Trade-off relation between different objective functions is another format to interpret the DM's preference information. Yang et al. proposed a series of methods to search for the SOI by taking the indifference trade-offs elicited by the DM as the preference model~\cite{Yang99,YangL02,LuqueYW09}. They proposed the GRIST method to estimate the gradient of the utility function and project the gradient onto the tangent hyperplane of the PF. To promote the GRIST method for solving problems without nice mathematical properties such as   convexity and differentiability, Chen et al.~\cite{ChenXC17} proposed to use an EA to replace the gradient descent method.

\subsubsection{Polarized solution(s) selection}

Asking the DM to periodically select the most preferred and/or the most dislike solution(s) from a set of candidates is another alternative way to represent the DM's preference. For example, Folwer et al.~\cite{FowlerGKKMW10} proposed a cone dominance relation based on convex preference cones by asking the DM to specify the best and the worst solutions from the current population. Sinha et al.~\cite{SinhaKWD10} proposed a modified Pareto dominance relation based on polyhedral cones. It is built by asking the DM to select the most preferred solution(s) from an archive. By considering the uncertainty associated with the interactive EMO, Gong et al.~\cite{GongSJ13} proposed to convert the uncertainty into interval parameters and developed a preference polyhedron, constructed by convex cones, to approximate the DM's preference. In~\cite{KoksalanK10}, K\"oksalan and Karahan proposed an interactive version of territory defining EA~\cite{KarahanK10} where a territory is defined around each solution and the favourable weights of the best solution selected by the DM are identified to determine a new preferred weight region. In~\cite{GongLZJZ11}, Gong et al. proposed an interactive MOEA/D that uses the current best solution as an anchor to update the distribution of weight vectors to the ROI.

\subsubsection{Performance score}

Another natural way to express the preference is scoring within a given range of numeric numbers. Recently, the first author and his collaborators~\cite{LiCSY19} proposed an interactive EMO framework specifically designed for the decomposition multi-objective optimization. It periodically invites the DM to assign scores over some selected solutions according to their satisfaction to the DM's preference. Based on the collected scoring results, a radial basis function network is applied to build the preference model and it is used as the fitness function to guide the population towards the SOI in the next several iterations. In particular, the preferred search directions are expressed as a set of weight vectors biased towards the potential SOI.

\begin{remark}
    According to the above brief literature review and the summary in~\pref{tab:literature}, we can see that existing interactive EMO approaches are algorithm-specific. In other words, they are specifically designed upon a baseline algorithm, e.g., NSGA-II, IBEA or MOEA/D. There is no solution applicable for all Pareto-, indicator-, and decomposition-based EMO frameworks. Note that our recent study~\cite{LiDAY17,LiLDMY20} has shown that all EMO frameworks are well suitable as a baseline for designing effective preference-based EMO algorithms.
\end{remark}


\begin{remark}
    Most, if not all, works use utility function as the preference model, the fitting of which is implemented as either a mathematical programming problem, a regression analysis or a fuzzy logic. As a subfield in machine learning, LTR, as known as machine learned ranking, is a powerful tool for preference learning from indirect information. It has been widely studied in information retrieval~\cite{Liu09} and recommendation systems~\cite{GemmisILMNS10}. However, it has rarely been considered in the context of interactive EMO, except~\cite{BattitiP10} to the best of our knowledge, which applied a support vector machine to serve the preference learning purpose.
\end{remark}

\begin{remark}
    Note that, before the development of interactive EMO, there have been a plethora of studies on interactive EAs that optimize systems based on subjective human evaluation~\cite{Takagi01}. Since they are mainly about single-objective optimization, of which the fitness function is determined without trading off conflicting objective functions, they are not directly applicable and out of the context of this paper.
\end{remark}


\section{Proposed Method}
\label{sec:method}

Our proposed interactive EMO framework based on a LTR neural network is a closed-loop of three modules including \texttt{consultation}, \texttt{preference elicitation} and \texttt{optimization}. Its termination is either called out by the DM or the exhaustion of the computational budget. In the following paragraphs, we will delineate the implementation of each module step by step.


\subsection{Consultation Module}
\label{sec:consultation}

The \texttt{consultation} module is the interface where the DM interacts with the EMO algorithm. The DM is asked to specify her preference over a set of selected candidate solutions $\mathcal{S}=\{\tilde{\mathbf{x}}^i\}_{i=1}^\mu$, $1\leq\mu\ll N$. Then, a preference model is learned based on the collected preference information. To this end, we need to address the following three core questions.

\subsubsection{Which solutions are chosen for consultation}

As discussed in~\cite{LiCSY19}, it is arguable to simply ask the DM to compare all solutions in a population every generation. This makes the search be completely driven by the DM. Thus, it significantly increases her cognitive load and is highly likely to lead to a fatigue. As discussed in~\cite{BattitiP10}, the DM can hardly make reasonable judgements on poorly converged solutions, thus it might not be helpful or even detrimental to consult the DM at the early stage of evolution. In this paper, we fix the number of consultations, say every $\tau>1$ generations, after running an EMO algorithm without considering any DM's preference information for several generations. During each consultation session, only a limited number of $\mu$ incumbent solutions, evaluated by the utility function learned by our preference model introduced in~\pref{sec:preference_model}, are chosen to constitute $\mathcal{S}$ for preference elicitation.

\subsubsection{What preference information do we ask the DM to give}

As overviewed in~\pref{sec:related}, there are multiple ways for the DM to specify her preference information. In this paper, we consider the holistic pairwise comparisons as the indirect preference information. Specifically, during the consultation stage, the DM is asked to iteratively decide, according to her preference, the quality of a solution pair $\langle\tilde{\mathbf{x}}^i,\tilde{\mathbf{x}}^j\rangle$ chosen from $\mathcal{S}$ where $i,j\in\{1,\cdots,\mu\}$ and $i\neq j$. The outcome of each pairwise comparison is either $\tilde{\mathbf{x}}^i$ is better, worse or indifferent over $\tilde{\mathbf{x}}^j$, denoted as $\tilde{\mathbf{x}}^i\succ\tilde{\mathbf{x}}^j$, $\tilde{\mathbf{x}}^i\prec\tilde{\mathbf{x}}^j$ or $\tilde{\mathbf{x}}^i\simeq\tilde{\mathbf{x}}^j$. In total, there are $\binom{\mu}{2}$ pairwise comparisons thus leading to $\binom{\mu}{2}$ holistic indirect judgements.

\subsubsection{How to learn a preference model}
\label{sec:preference_model}

Based on the collected preference information, i.e., the holistic indirect judgements, the goal of preference learning is to learn a preference model that is able to evaluate the quality of solutions according to the DM's preference. In principle, this model is a utility function $u(\mathbf{x}): \mathbb{R}^m\rightarrow\mathbb{R}$ such that the ranking order of a set of testing samples satisfy that $u(\mathbf{x}^i)>u(\mathbf{x}^j)$ if $\mathbf{x}^i\succ\mathbf{x}^j$ and $u(\mathbf{x}^i)=u(\mathbf{x}^j)$ if $\mathbf{x}^i\simeq\mathbf{x}^j$.

\begin{figure}[t!]
\centering
\includegraphics[width=.6\linewidth]{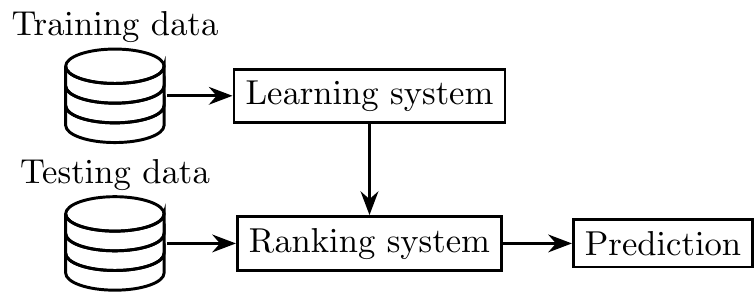}
\caption{A flowchart of a classic LTR routine.}
\label{fig:LTR}
\end{figure}

LTR is a machine learning approach, typically as a supervised or a semi-supervised learning, to construct a ranking model. LTR has been a core part of modern information retrieval systems such as document retrieval~\cite{DeveaudMUN19}, collaborative filtering~\cite{PessiotTUAG07}, sentiment analysis~\cite{Liu11} and online advertising~\cite{KarimzadehganLZM11}. In practice, the training data of a LTR model consist of lists of items with some partial orders specified by the user queries between items in each list. This order is typically induced by giving a numerical or ordinal score or a binary judgment, such as \lq relevant\rq\ versus \lq irrelevant\rq\ for each item, also known as point-, list-, and pairwise query respectively. A LTR model aims to predict the ranking order of a permutation of items in new and unseen dataset. \pref{fig:LTR} gives a basic workflow of a classic LTR routine.

According to the above description, we appreciate that LTR shares a common philosophy as our preference learning purpose. In this paper, we propose to use LTR as an alternative preference model. The quality of the underlying preference model is measured by the relative comparison results between pairs of candidate solutions. In other words, the less inaccurate order of solution pairs found, the better preference model is. Let $\mathbb{P}(\mathbf{x}^i\succ\mathbf{x}^j)$ denote as the posterior probability predicted by our preference model that conforms $\mathbf{x}^i\succ\mathbf{x}^j$ and it is defined as a variant of sigmoid function:
\begin{equation}
    \mathbb{P}(\mathbf{x}^i\succ\mathbf{x}^j)=\frac{1}{1+e^{-\sigma(u(\mathbf{x}^i)-u(\mathbf{x}^j))}},
    \label{eq:predicted_prob}
\end{equation}
where $\sigma>0$ is a control parameters that determines the shape of $\mathbb{P}(\mathbf{x}^i\succ\mathbf{x}^j)$. We denote $\widetilde{\mathbb{P}}(\mathbf{x}^i\succ\mathbf{x}^j)$ as the ground truth probability of $\mathbf{x}^i\succ\mathbf{x}^j$ and it is calculated as:
\begin{equation}
    \widetilde{\mathbb{P}}(\mathbf{x}^i\succ\mathbf{x}^j)=\frac{1}{2}(1+c_{ij}),
    \label{eq:true_prob}
\end{equation}
where $c_{ij}=1,-1,$ or $0$ indicates that $\mathbf{x}^i\succ\mathbf{x}^j$, $\mathbf{x}^j\succ\mathbf{x}^i$ or $\mathbf{x}^i\simeq\mathbf{x}^j$ respectively. Inspired by~\cite{BurgesSRLDHH05}, to improve the robustness of the ranking result, our preference model training uses the cross-entropy between $\mathbb{P}(\mathbf{x}^i\succ\mathbf{x}^j)$ and $\widetilde{\mathbb{P}}(\mathbf{x}^i\succ\mathbf{x}^j)$ as the loss function associated with the ranking pair $\langle\mathbf{x}^i,\mathbf{x}^j\rangle$:
\begin{align}
    \ell_{ij}&=-\widetilde{\mathbb{P}}(\mathbf{x}^i\succ\mathbf{x}^j)\log\mathbb{P}(\mathbf{x}^i\succ\mathbf{x}^j)\nonumber\\
    &-(1-\widetilde{\mathbb{P}}(\mathbf{x}^i\succ\mathbf{x}^j))\log(1-\mathbb{P}(\mathbf{x}^i\succ\mathbf{x}^j)).
\end{align}
By using equations~(\ref{eq:predicted_prob}) and (\ref{eq:true_prob}), $\ell_{ij}$ is simplified as:
\begin{align}
    \ell_{ij}&=\frac{1}{2}(1-c_{ij})\sigma(u(\mathbf{x}^i)-u(\mathbf{x}^j))\nonumber\\
    &+\log(1+e^{-\sigma(u(\mathbf{x}^i)-u(\mathbf{x}^j))}),
\end{align}
where
\begin{equation}
    \ell_{ij}=
    \begin{cases}
        \log(1+e^{\sigma(u(\mathbf{x}^j)-u(\mathbf{x}^i))}), & \text{if}\ c_{ij}=1\\
        \log(1+e^{\sigma(u(\mathbf{x}^i)-u(\mathbf{x}^j))}), & \text{if}\ c_{ij}=-1
    \end{cases}.
\end{equation}
Note that there are two characteristics of this loss function.
\begin{itemize}
    \item If $u(\mathbf{x}^i)$ equals $u(\mathbf{x}^j)$ but $\mathbf{x}^i\simeq\mathbf{x}^j$ does not conform, we still have $\ell_{ij}>0$ that penalizes this pair $\langle\mathbf{x}^i,\mathbf{x}^j\rangle$ thus leading to their ranking apart from each other.
    \item The overall loss function is the summation of the loss function of each ranking pair:
        \begin{equation}
            \mathcal{L}=\sum_{\langle\mathbf{x}^i,\mathbf{x}^j\rangle\in I}\ell_{ij},
        \end{equation}
        where $I$ is the set of all ranking pairs and this loss function asymptotes to a linear function.
\end{itemize}
In this paper, we apply a neural network with a single hidden layer to learn the preference model. Since $\mathcal{L}$ is differentiable, stochastic gradient descent (SGD)~\cite{Bottou12} is used to update the weights $\tilde{\mathbf{w}}=(\tilde{w}_1,\cdots,\tilde{w}_\ell)^T$ of the neural network:
\begin{equation}
    \tilde{w}^{\prime}_k=\tilde{w}_k-\eta\frac{\partial\mathcal{L}}{\partial \tilde{w}_k},
\end{equation}
where $k\in\{1,\cdots,\ell\}$ and $\eta>0$ is the learning rate. If the update of $\mathcal{L}$ is along the opposite direction of the gradient:
\begin{align}
    \Delta\mathcal{L}&=\sum_{k=1}^\ell\frac{\partial\mathcal{L}}{\partial \tilde{w}_k}\Delta \tilde{w}_k=\sum_{k=1}^\ell\frac{\partial\mathcal{L}}{\partial \tilde{w}_k}(-\eta\frac{\partial\mathcal{L}}{\partial \tilde{w}_k})\nonumber\\
    &=-\eta\sum_{k=1}^\ell(\frac{\partial\mathcal{L}}{\partial \tilde{w}_k})^2<0,
\end{align}
thereby reducing $\mathcal{L}$. If we further decompose the gradient term as follows:
\begin{equation}
    \frac{\partial\mathcal{L}}{\partial \tilde{w}_k}=\sum_{\langle\mathbf{x}^i,\mathbf{x}^j\rangle\in I}(\frac{\partial\mathcal{L}_{ij}}{\partial u(\mathbf{x}^i)}\frac{\partial u(\mathbf{x}^i)}{\partial \tilde{w}_k}+\frac{\partial\mathcal{L}_{ij}}{\partial u(\mathbf{x}^j)}\frac{\partial u(\mathbf{x}^j)}{\partial \tilde{w}_k}),
    \label{eq:gradient}
\end{equation}
where
\begin{align}
    \frac{\partial\mathcal{L}_{ij}}{\partial u(\mathbf{x}^i)}&=\sigma(\frac{1}{2}(1-c_{ij})-\frac{1}{1+e^{\sigma(u(\mathbf{x}^i)-u(\mathbf{x}^j))}})\nonumber\\
    &=-\frac{\partial\mathcal{L}_{ij}}{\partial u(\mathbf{x}^j)}.
\end{align}
Let $\lambda_{ij}=\frac{\partial\mathcal{L}_{ij}}{\partial u(\mathbf{x}^i)}$, then we can rewrite \pref{eq:gradient} as:
\begin{equation}
    \frac{\partial\mathcal{L}}{\partial \tilde{w}_k}
    =\sum_{\langle\mathbf{x}^i,\mathbf{x}^j\rangle\in I}\lambda_{ij}(\frac{\partial u(\mathbf{x}^i)}{\partial \tilde{w}_k}-\frac{\partial u(\mathbf{x}^j)}{\partial \tilde{w}_k}).
\end{equation}
This formulation makes the training of our neural network as a mini-batch learning process which reduces the computational complexity of SGD from quadratic to linear.

\subsubsection{Working example}
\label{sec:example}

Here we use a simple example to illustrate the learning process of our preference model. For simplicity, let us assume that the unknown utility function standing for the DM's preference is the Tchebycheff function:
\begin{equation}
g(\mathbf{x}|\mathbf{w},\mathbf{z}^\ast)=\max \limits_{1\leq i \leq m} |f_i(\mathbf{x})-z^\ast_i|/w_i,
\end{equation}
where we assume that $\mathbf{z}^\ast=\mathbf{0}$ and $w_i>0$. We use the 3-objective DTLZ1~\cite{DebTLZ05} as an example and the middle of its PF is assumed to be preferred by the DM, with $\mathbf{w}=(1,1,1)^T$. Given four solutions $\{\mathbf{x}^i\}_{i=1}^4$ as listed below:
\begin{center}
\begin{tabular}{c|c|c|c|c|c}
   & $f_1(\mathbf{x})$ & $f_2(\mathbf{x})$ & $f_3(\mathbf{x})$ & $g(\mathbf{x}|\boldsymbol w)$ & $r(\mathbf{x})$ \\\hline
   $\mathbf{x}^1$ & $0.167$ & $0.167$ & $0.167$ & $0.167$  & $1$ \\ \hline
   $\mathbf{x}^2$ & $0.2$ & $0.15$ & $0.15$ & $0.2$  & $2$ \\ \hline
   $\mathbf{x}^3$ & $0.3$ & $0.1$ & $0.1$ & $0.3$  & $3$ \\ \hline
   $\mathbf{x}^4$ & $0.4$ & $0.05$ & $0.05$ & $0.4$ & $4$ \\
\end{tabular}
\end{center}
where $r(\mathbf{x})\in\{1,\cdots,4\}$ is the preference rank of a solution. In particular, the smaller the $r(\mathbf{x})$ is, the more preferred $\mathbf{x}$ is. We can have $\binom{4}{2}$ $=12$ pairwise comparisons in total while we only choose three pairs, i.e., $\mathbf{x}^1\succ\mathbf{x}^2, \mathbf{x}^1\succ\mathbf{x}^3, \mathbf{x}^2\succ\mathbf{x}^4$ to train our neural network.  

\begin{figure}[t!]
\centering
\includegraphics[width=.6\linewidth]{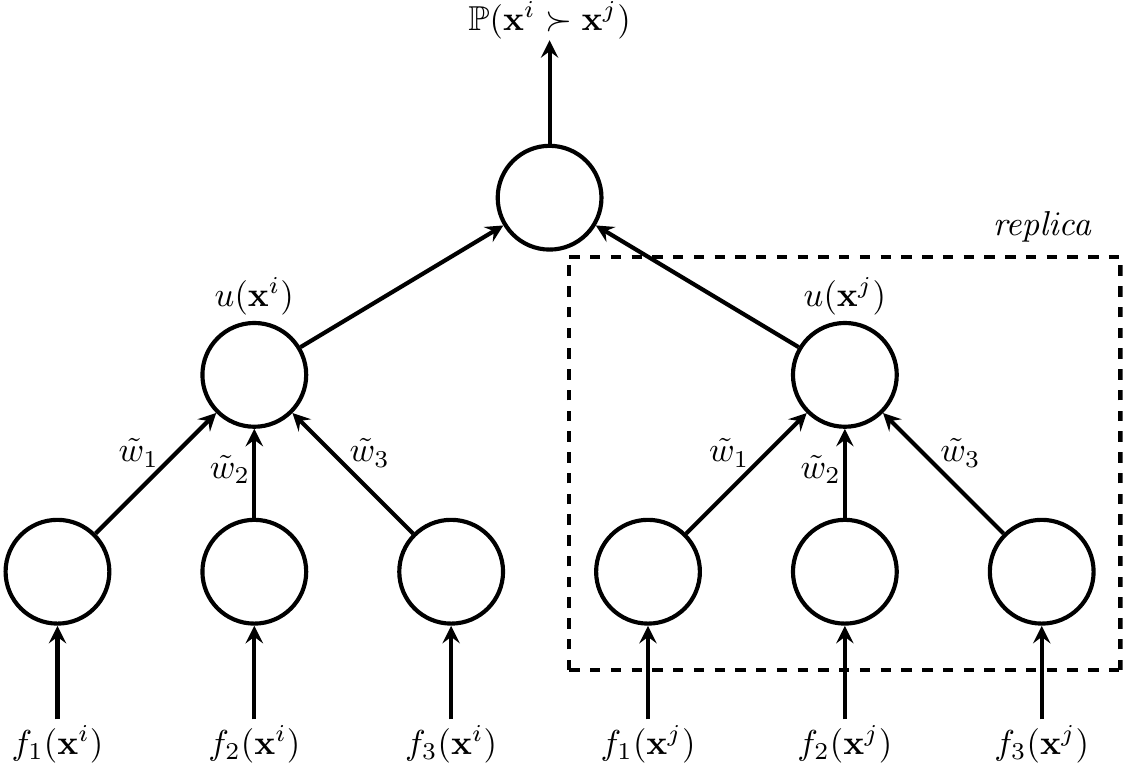}
\caption{An illustrative example of the working mechanism of our neural network based LTR model.}
\label{fig:working_example}
\end{figure}

We use a simple neural network without any hidden layer, shown in \pref{fig:working_example}, as an example to explain the working mechanism of our preference model.

In this case, the output of our preference model is a weighted aggregation as  $u(\mathbf{x})=\sum_{i=1}^m\tilde{w}_if_i(\mathbf{x})$ where $\tilde{w}_i$ is a weight that needs to be learned in our neural network training process. For example, let us assume that the initial weights are set as $\tilde{\mathbf{w}}=(0.620,-0.952,-0.734)^T$. After the training process, we have $\tilde{\mathbf{w}}^\ast=(-2.846,2.524,2.742)^T$, and the corresponding utility function values predicted by our preference model are listed below:

\begin{center}
\begin{tabular}{c|c|c|c|c|c}
   & $f_1(\mathbf{x})$ & $f_2(\mathbf{x})$ & $f_3(\mathbf{x})$ & $u^0(\mathbf{x})$ & $u^\ast(\mathbf{x})$ \\\hline
   $\mathbf{x}^1$ & $0.167$ & $0.167$ & $0.167$ & $-0.178$  & $0.404$ \\ \hline
   $\mathbf{x}^2$ & $0.2$ & $0.15$ & $0.15$ & $-0.129$  & $0.221$ \\ \hline
   $\mathbf{x}^3$ & $0.3$ & $0.1$ & $0.1$ & $0.017$  & $-0.327$ \\ \hline
   $\mathbf{x}^4$ & $0.4$ & $0.05$ & $0.05$ & $0.164$ & $-0.875$ \\
\end{tabular}
\end{center}
where $u^0(\mathbf{x})$ and $u^\ast(\mathbf{x})$ respectively indicates the preference scores assigned to solutions by the preference model before and after the neural network training. Note that the ground truth utility function values of $\{\mathbf{x}_i\}_{i=1}^4$ are unknown a priori when training the preference model. From this example, we can see that the rank inferred from the predicted preference scores conforms to the ground truth.

\subsection{Preference Elicitation Module}
\label{sec:elicition}

The \texttt{preference elicitation} module transforms the preference information learned in the \texttt{consultation} module into the format that can be used in the \texttt{optimization} module. There are three major EMO frameworks, i.e., Pareto-, indicator- and decomposition-based, in the literature. Each one differs from the others mainly in the environmental selection, i.e., the way of survival the fittest. In this subsection, we choose three iconic EMO algorithms from these three frameworks as the baseline and tailor our preference model to their environmental selection.

\subsubsection{Pareto-based EMO algorithm}
\label{sec:pareto}

It usually consists of two parts: one is the use of Pareto dominance to push the population towards the PF (i.e., convergence) and the other is the use of a density estimation metric to maintain the population diversity. Here we choose NSGA-II, which has been widely recognized as one of the most successful algorithms, as the baseline for the Pareto-based EMO algorithm~\cite{LiKWCR12,LiKD15,LiDZ15}. In practice, we keep the fast non-dominated sorting~\cite{LiDZZ17} untouched since the Pareto optimality is always the first priority in multi-objective optimization. The utility function learned by our preference model in the \texttt{consultation} module is used as the alternative of crowding distance in NSGA-II.

\subsubsection{Decomposition-based EMO algorithm}
\label{sec:decomposition}

\begin{figure}[htbp]
\centering
\subfloat[Original distribution.]{\includegraphics[width=.3\linewidth]{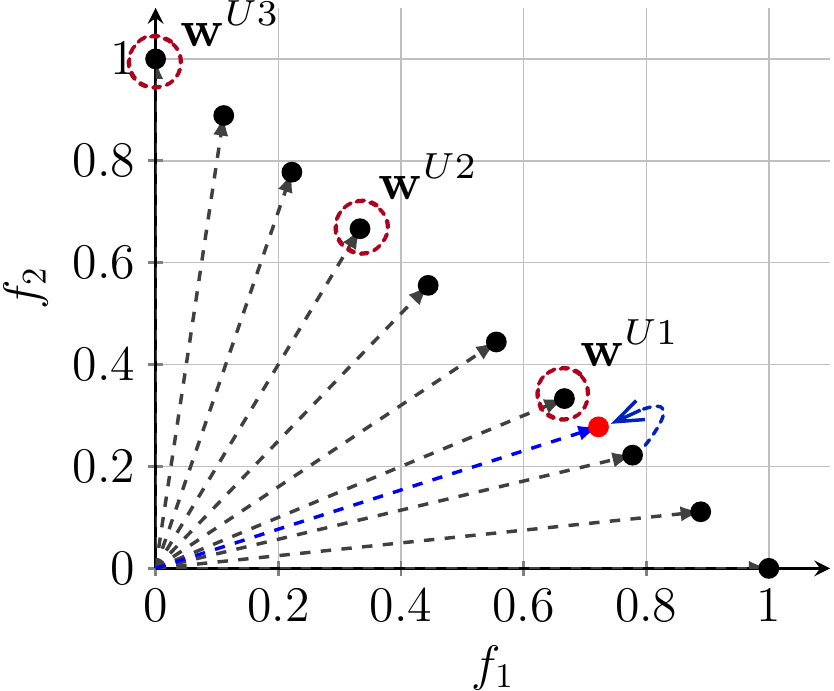}}
\subfloat[Adjusted distribution.]{\includegraphics[width=.3\linewidth]{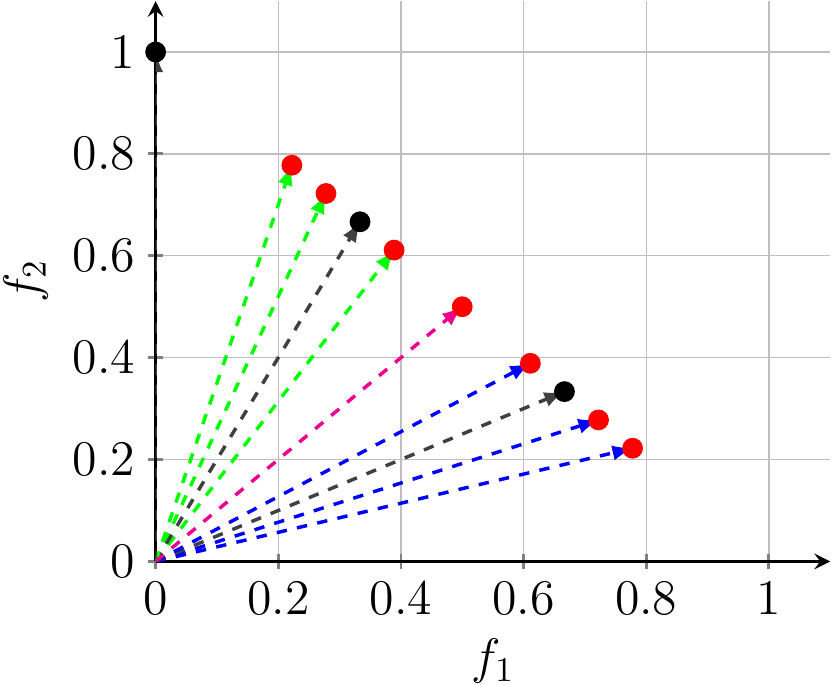}}
\caption{Illustration of the preference elicitation process.}
\label{fig:pref_weight}
\end{figure}

Here we focus on MOEA/D which has been widely recognized as the iconic decomposition-based EMO algorithm. Its basic idea is to decompose the original MOP into a set of subproblems, either as scalarizing functions or simplified MOPs. Thereafter, a population-based meta-heuristic is applied to solve these subproblems in a collaborative manner. In MOEA/D, the weight vectors used to define the subproblems can be regarded as the driver to represent the DM's preference information~\cite{WuKZLWL15,WuKJLZ17,WuLKZZ19,WuLKZ20}. In particular, each weight vector indicates a preferred region on the PF. The \texttt{preference elicitation} module mainly aims to change the originally uniformly distributed weight vectors $\mathcal{W}=\{\mathbf{w}^i\}_{i=1}^N$ to be biased towards the region of interest. Here we follow the same four-step procedure developed in our recent work~\cite{LiCSY19} to serve this purpose.
\begin{enumerate}[Step 1:]
    \item Use $u(\mathbf{x})$ learned in the consultation module to score each member of the current population $\mathcal{P}$.
    \item Rank the population according to the scores assigned in Step 1, and find the top $\mu$ solutions. Weight vectors associated with these solutions are deemed as the promising ones, and store them in a temporary archive $\mathcal{W}^U=\{\mathbf{w}^{Ui}\}_{i=1}^{\mu'}$ where $1\leq\mu'\leq\mu$.
    \item For $i=1$ to $\mu'$ do 
        \begin{enumerate}[Step 3.1:]
            \item Find the $\lceil\frac{N-\mu'}{\mu'}\rceil$ closest weight vectors to $\mathbf{w}^{Ui}$ according to their Euclidean distances.
            
            \item Move each of these weight vectors towards $\mathbf{w}^{Ui}$ as follows:
            \begin{equation}
                 w_j=w_j+\eta\times(\mathbf{w}^{Ui}_j-w_j),
                \label{eq:weight_update}
            \end{equation}
            where $j\in\{1,\cdots,m\}$.

            \item Temporarily remove these weight vectors from $\mathcal{W}$ and go to Step 3.
        \end{enumerate}
    \item Output the adjusted reference points as the new $\mathcal{W}$.
\end{enumerate}
In particular, $0<\eta\leq 1$ is the step size used to tweak the weight vectors.~\pref{fig:pref_weight} gives an example of this preference elicitation process in a two-objective case. Three promising reference points are highlighted by red circles. $\mathbf{w}^{U1}$ has the highest priority to attract its companions, and so on. Interested readers are referred to~\cite{LiCSY19} for more detail.

\begin{figure}[htbp]
\centering
\includegraphics[width=.6\linewidth]{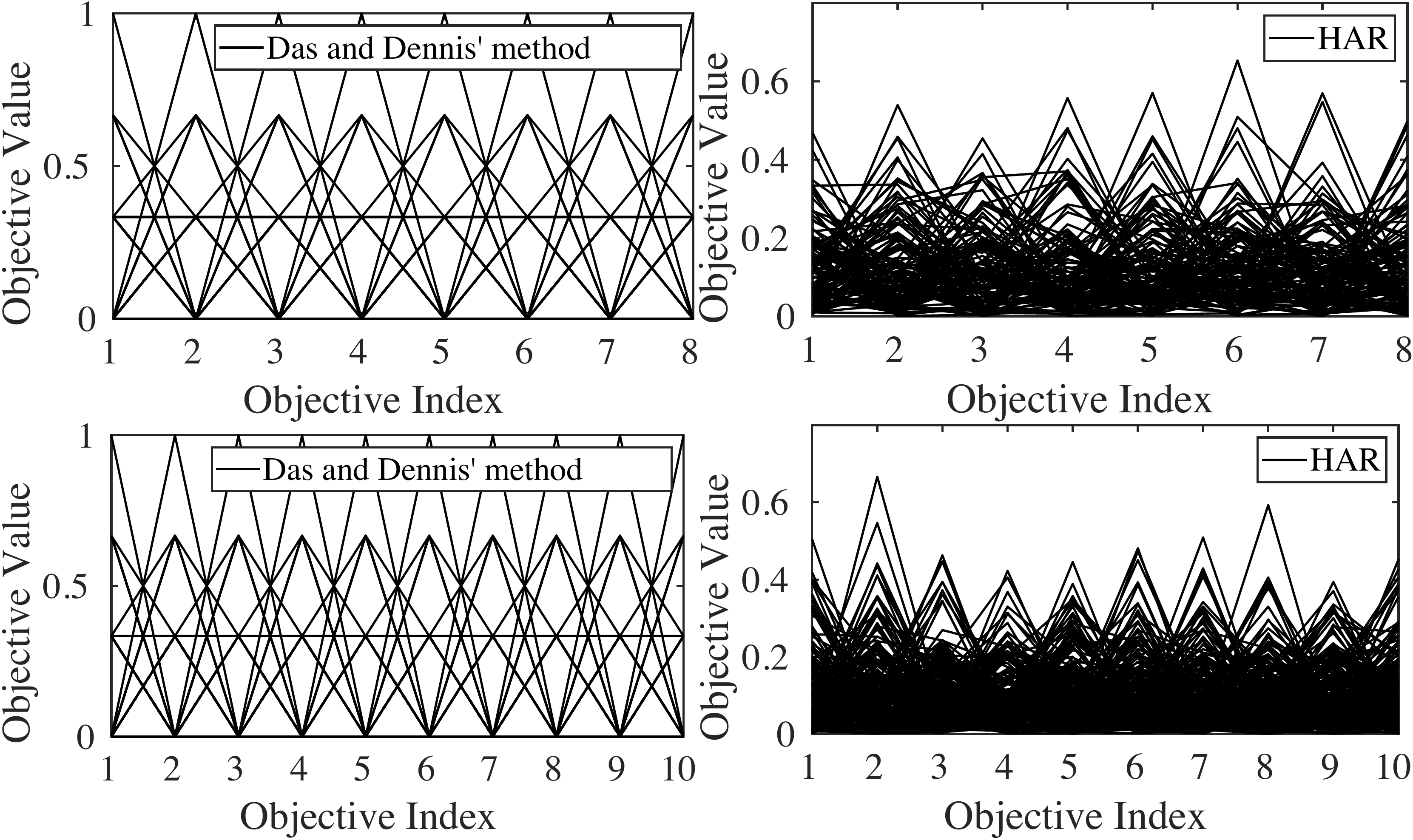}
\caption{Comparisons of the $120$ and the $220$ weight vectors generated by using the Das and Dennis' method~\cite{DasD98} versus the HAR method~\cite{CiomekK21} on $8$ and $10$ objectives, repsepctively.}
\label{fig:NBI_vs_HAR}
\end{figure}

\begin{remark}
    From our preliminary experiments, we find that the distribution of the initial weight vectors has a significant impact on the weight vector adjustment in Step 3. This can be attributed to the sparse distribution of weight vectors generated by the widely used Das and Dennis' method~\cite{DasD98}. As the $8$- and $10$-objective examples shown in~\pref{fig:NBI_vs_HAR}, we can see that almost all weight vectors are sparsely distributed along the boundary of the simplex while there are few lying in the middle part. This renders the weight adjustment in Step 3 hardly effective in a high-dimensional scenario. To mitigate this issue, inspired by~\cite{TomczykK20}, we apply the hit-and-run (HAR) method~\cite{CiomekK21} as an alternative of the Das and Dennis' method for the initialization of weight vectors. Note that the HAR method is a variant of Markov Chain Monte Carlo and is scalable to a high-dimensional space~\cite{TervonenVBP13}. Its basic idea is to sample as a set of uniformly distributed weight vectors from a constrained simplex. Specifically, for each weight vector $\mathbf{w}^i\in\mathcal{W}$, $i\in\{1,\cdots,N\}$, it should satisfy some linear constraints $\sum_{j=1}^mw_j^i=1$ and $w_j^L\leq w_j^i\leq w_j^U$ where $w_j^L$ and $w_j^U$ are the lower and upper bounds for the $j$-th component of the weight vector $\mathbf{w}^i$. In particular, we set $w_j^L=0$ and $w_j^U=1$ for $j\in\{1,\cdots,m\}$ in this paper. From the $8$- and $10$-objective examples shown in~\pref{fig:NBI_vs_HAR}, we can see that weight vectors generated by the HAR method have a descent distribution in the intermediate section of the simplex.
\end{remark}

\subsubsection{Indicator-based EMO algorithm}
\label{Indicator-based}

Its basic idea is to apply a performance indicator to transform a MOP into a single-objective optimization problem. Instead of using the binary $\epsilon$-indicator as in the classic IBEA, we opt to the $R2$ indicator~\cite{HansenJ87} in view of its encouraging results reported for multi- and many-objective optimization~\cite{ShangIN20} as well as preference articulation~\cite{WagnerTB13}.

Specifically, $R2$ indicator is used to evaluate the relative quality of two sets of trade-off solutions by using the weighted Tchebycheff function with a given reference point $\mathbf{z}=(z_1,\cdots,z_m)^T$.
\begin{equation}
R2(\mathcal{P},\mathcal{W},\mathbf{z})=\sum_{i=1}^m\bigg(\mathbb{P}(\mathcal{W})\times\min_{\mathbf{x}^i\in\mathcal{P}}\Big\{\max_{1\leq j\leq m}w^i|x^i_j-z_j|\Big\}\bigg),
\label{R2-indicator_1}
\end{equation}
where $\mathcal{P}$ is the current population, $\mathcal{W}$ is a set of weight vectors and $\mathbb{P}(\mathcal{W})$ indicates the probability distribution on $\mathcal{W}$. In particular, the $R2$ indicator can be rewritten as follows when the weight vectors are evenly distributed:
\begin{equation}
R2(\mathcal{P},\mathcal{W},\mathbf{z})=\frac{1}{|\mathcal{W}|}\sum_{\mathbf{w}^i\in \mathcal{W}}\bigg(\min_{\mathbf{x}^i\in\mathcal{P}}\Big\{\max_{1\leq j\leq m}w^i|x^i_j-z_j|\Big\}\bigg).
\label{eq:R2}
\end{equation}
In this case, we can see that the DM's preference information can also be represented as a set of biased weight vectors. In other words, we use a set of adjusted weight vectors obtained by the four-step procedure introduced in~\pref{sec:decomposition} to replace the $\mathcal{W}$ in~\pref{eq:R2}.

\subsection{Optimization Module}
\label{Optimization}

In principle, any EMO algorithm can be adapted to be a baseline algorithm in the \texttt{optimization} module. This paper chooses NSGA-II, MOEA/D and R2-IBEA, as discussed in~\pref{sec:elicition} for proof-of-concept purposes, to generate three algorithm instances of our proposed framework, denoted as \texttt{I-NSGA-II/LTR}, \texttt{I-MOEA/D/LTR} and \texttt{I-R2-IBEA/LTR} respectively. Note that all these algorithm instances run as the vanilla version without considering the DM's preference before the first consultation session.



\section{Experimental Settings}
\label{sec:settings}

This section introduces the experimental settings of the empirical studies in this paper, including the benchmark problems, parameter settings, peer algorithms, performance metrics and statistical tests~\cite{ChenLY18,ZouJYZZL19,LiZZL09,BillingsleyLMMG19,LiZLZL09,LiFK11,LiKWTM13,CaoKWL12,CaoKWL14,LiWKC13,CaoKWLLK15,LiXT19,LiuLC19,LiZ19,KumarBCLB18,CaoWKL11,LiX0WT20,LiuLC20,LiXCT20,WangYLK21,ShanL21,LiLLM21}.

\subsection{Benchmark Problem Suite}
\label{sec:benchmark}

In this paper, we consider test problems chosen from five widely used benchmark suites including, DTLZ1 to DTLZ6~\cite{DebTLZ05}, DTLZ1$^{-1}$ to DTLZ4$^{-1}$~\cite{IshibuchiSMN17}, mDTLZ1 to mDTLZ4~\cite{WangOI19} and WFG3~\cite{IshibuchiMN16}. All these test problems are with continuous variables and have various PF shapes (e.g., linear, convex, concave, disconnected, degenerate and inverted PFs) and different search space properties. In our experiments, we consider $m\in\{3,5,8,10\}$ except the mDTLZ problems which are constantly with three objectives. The number of variables are set as recommended in their original papers.

\subsection{Parameter Settings}
\label{sec:parameters}

The parameters associated with our proposed interactive EMO algorithms are outlined as follows.
\begin{itemize}
\item The number of incumbent candidates presented to the DM for pairwise comparisons: $\mu=10$;
\item The number of generations between two consecutive consultation sessions: $\tau=10$;
\item The step size of the reference point update used in~\pref{eq:weight_update}: $\eta=0.2$;
\item The number of function evaluations (FEs) and population size  settings are given in Table 9 of the supplemental material\footnote{The supplemental materials can be found in \url{https://tinyurl.com/2p962bpc}.} as suggested in~\cite{TomczykK20};
\item The crossover probability and the distribution index for the simulated binary crossover operator~\cite{DebA94}: $p_c=1.0$ and $\eta_c=30$;
\item The mutation probability and the distribution index for the polynomial mutation operator~\cite{DebA99}: $p_c=1/n$ and $\eta_m=20$;
\item The control parameter of the sigmoid function: $\sigma=1$;
\end{itemize}

\subsection{Peer Algorithms}
\label{sec:peer_algorithms}

To validate the competitiveness of the proposed interactive EMO algorithms, we compare their performance with four SOTA peer algorithms in the literature, i.e., \texttt{BC-EMO}~\cite{BattitiP10}, \texttt{NEMO-0}~\cite{BrankeGSZ15}, \texttt{I-MOEA/D-PLVF}~\cite{LiCSY19}, and \texttt{IEMO/D}~\cite{TomczykK20}. Note that \texttt{BC-EMO} is the only peer algorithm, to the best of our knowledge, that applies LTR to serve the preference learning purpose. Although the representation of the DM's preference information in \texttt{I-MOEA/D-PLVF} is different from this paper, it is still worthwhile to be compared since it is the first instantiation of the interactive EMO framework shown in~\pref{fig:flowchart}. The corresponding parameters are set according to the recommendations in their original papers. Interested readers are referred to the corresponding papers for more technical details of these peer algorithms. 

\subsection{Performance Evaluation}
\label{sec:metrics}

\subsubsection{Performance Metrics}

As discussed in our recent paper~\cite{LiDY18}, performance evaluation of interactive EMO methods is far from trivial as the choice of DM model can lead to potential bias. In this paper, we first consider a prescribed golden value function, which is unknown to an interactive EMO algorithm, to play as the artificial DM:
        \begin{equation}
            \psi(\mathbf{x})=\max_{1\leq i\leq m}|f_i(\mathbf{x})-z_i^\ast|/w^{\ast}_i,
            \label{eq:vf}
        \end{equation}
        where $\mathbf{z}^\ast=(z_1,\cdots,z_m)^T$ is set to be the origin in our experiments, and $\mathbf{w}^\ast=(w_1^\ast,\cdots,w_m^\ast)^T$ is the utopia weight that represents the DM's expected importance of different objectives. In this paper, we consider two types of $\mathbf{w}^\ast$: one prefers the solution with an equal importance priority over all objectives (denoted as $\mathbf{w}^{\mathtt{e}}$) while the other one prefers the solution with a focused priority over a particular objective (denoted as $\mathbf{w}^{\mathtt{b}}$), i.e., biased toward a particular side of the PF. Since a $m$-objective problem has $m$ sides, there can be $m$ different choices for setting the biased $\mathbf{w}^{\ast}$. In our experiments, we randomly choose one side for proof-of-concept purposes.

To evaluate the performance of an interactive EMO algorithm for approximating the ROI, we consider using the approximation error of the obtained population $\mathcal{P}$ w.r.t. the DM's golden point $\mathbf{x}^\ast$ (i.e., the Pareto-optimal solution for a given utopia weight) as the performance metric:
        \begin{equation}
            \mathbb{E}(\mathcal{P})=\min_{\mathbf{x}\in \mathcal{P}}\mathsf{dist}(\mathbf{x},\mathbf{x}^\ast),
        \end{equation}
        where $\mathsf{dist}(\mathbf{x},\mathbf{x}^\ast)$ is the Euclidean distance between $\mathbf{x}^\ast$ and a solution $\mathbf{x}\in \mathcal{P}$ in the objective space. The choice of $\mathbf{w}^\ast$ and $\mathbf{x}^\ast$ are listed in Tables 10 to 13 of the supplementary material but is unknown to the algorithm.

\subsubsection{Statistical Tests}
Each experiment is repeated independently $31$ times with different random seeds. To have a statistical interpretation of the significance of comparison results, three statistical measures are used in our empirical study.
\begin{itemize}
	\item\underline{Wilcoxon signed-rank test}~\cite{Wilcoxon1945IndividualCB}: This is a non-parametric statistical test that makes little assumption about the underlying distribution of the data and has been recommended in many empirical studies in the EA community~\cite{DerracGMH11}. In particular, the significance level is set to $p=0.05$ in our experiments.
    \item\underline{Scott-knott test}~\cite{MittasA13}: Instead of merely comparing the raw $\mathbb{E}(\mathcal{P})$ values, we apply the Scott-knott test to rank the performance of different peer techniques over $31$ runs on each experiment. In a nutshell, the Scott-knott test uses a statistical test and effect size to divide the performance of peer algorithms into several clusters. In particular, the performance of peer algorithms within the same cluster are statistically equivalent. Note that the clustering process terminates until no split can be made. Finally, each cluster can be assigned a rank according to the mean $\mathbb{E}(\mathcal{P})$ values achieved by the peer algorithms within the cluster. In particular, since a smaller $\mathbb{E}(\mathcal{P})$ value is preferred, the smaller the rank is, the better performance of the technique achieves.
    \item\underline{$A_{12}$ effect size}~\cite{VarghaD00}: To ensure the resulted differences are not generated from a trivial effect, we apply $A_{12}$ as the effect size measure to evaluate the probability that one algorithm is better than another. Specifically, given a pair of peer algorithms, $A_{12}=0.5$ means they are \textit{equivalent}. $A_{12}>0.5$ denotes that one is better for more than 50\% of the times. $0.56\leq A_{12}<0.64$ indicates a \textit{small} effect size while $0.64 \leq A_{12} < 0.71$ and $A_{12} \geq 0.71$ mean a \textit{medium} and a \textit{large} effect size, respectively. 
\end{itemize}
Note that both Wilcoxon signed-rank test and $A_{12}$ effect size are also used in the Scott-Knott test for generating clusters.


\section{Experimental Results and Discussions}
\label{sec:experiments}

We seek to answer the following research questions (RQs) through our empirical study in the following paragraphs.
\begin{itemize}
\item \textit{\underline{RQ1}: How are the performance comparisons among our proposed three algorithm instances?}
\item \textit{\underline{RQ2}: How is the performance of our proposed algorithm instances compared against the selected SOTA peers?}
\item \textit{\underline{RQ3}: How is the performance of our proposed LTR neural network compared against the LTR algorithms used in \texttt{NEMO-0} and \texttt{BC-EMO}?}
\item \textit{\underline{RQ4}: What is the impact of the hyper-parameters associated with our proposed framework?}
\item \textit{\underline{RQ5}: How is the robustness w.r.t. some random noise in DM's preference information?}
\end{itemize}

\subsection{Performance Comparison among Our Proposed Three Algorithm Instances}
\label{sec:our_algorithms_comparisons}

\begin{figure*}[htbp]
    \centering
    \includegraphics[width=\linewidth]{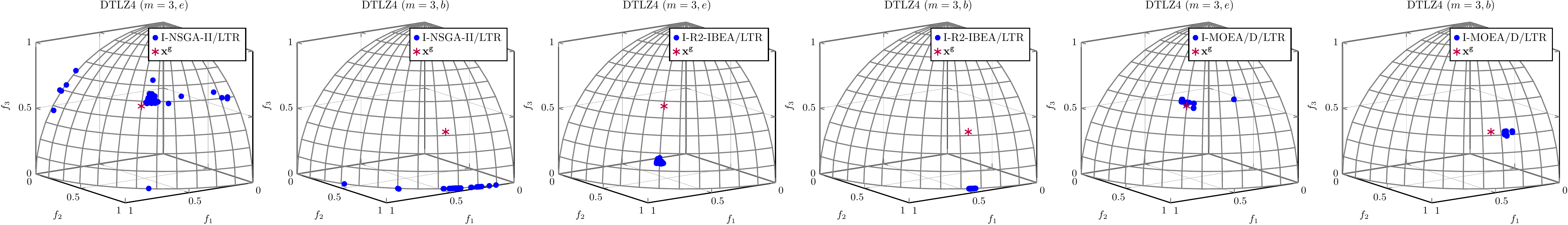}
    \caption{Population distribution of non-dominated solutions with the medium $\mathbb{E}(\mathcal{P})$ value obtained by \texttt{I-NSGA-II/LTR}, \texttt{I-R2-IBEA/LTR} and \texttt{I-MOEA/D/LTR} on DTLZ4 when $m=3$.}
    \label{fig:dtlz4_m3}
\end{figure*}

\begin{figure*}[htbp]
    \centering
    \includegraphics[width=\linewidth]{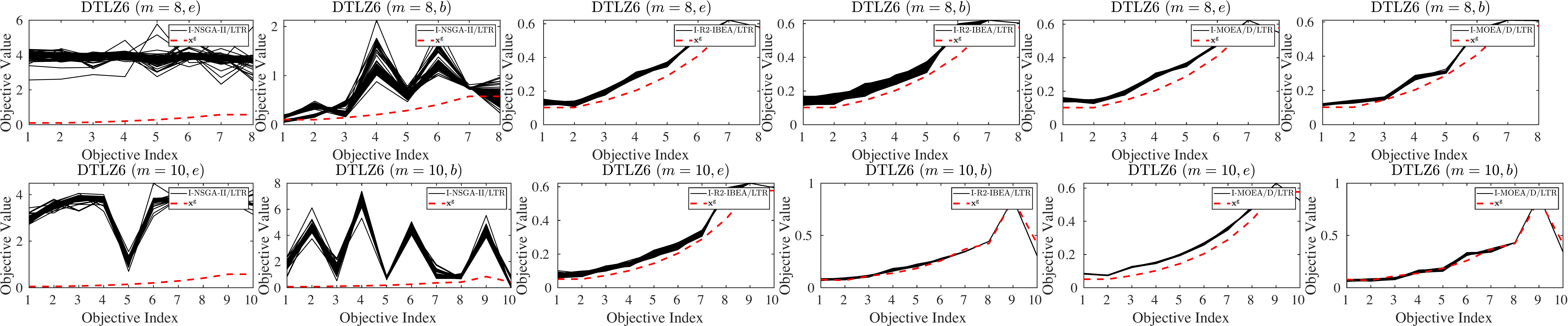}
    \caption{Population distribution of non-dominated solutions with the medium $\mathbb{E}(\mathcal{P})$ value obtained by \texttt{I-NSGA-II/LTR}, \texttt{I-R2-IBEA/LTR} and \texttt{I-MOEA/D/LTR} on DTLZ6 when $m=8$ and $m=10$, respectively.}
    \label{fig:dtlz6_m8_m10}
\end{figure*}

\begin{figure*}[htbp]
    \centering
    \includegraphics[width=\linewidth]{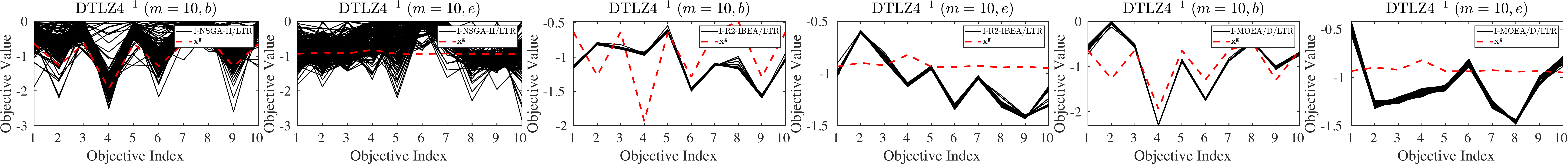}
    \caption{Population distribution of non-dominated solutions with the medium $\mathbb{E}(\mathcal{P})$ value obtained by \texttt{I-NSGA-II/LTR}, \texttt{I-R2-IBEA/LTR} and \texttt{I-MOEA/D/LTR} on DTLZ4$^{-1}$ when $m=10$.}
    \label{fig:minus_dtlz4_m10}
\end{figure*}

\begin{figure*}[htbp]
    \centering
    \includegraphics[width=\linewidth]{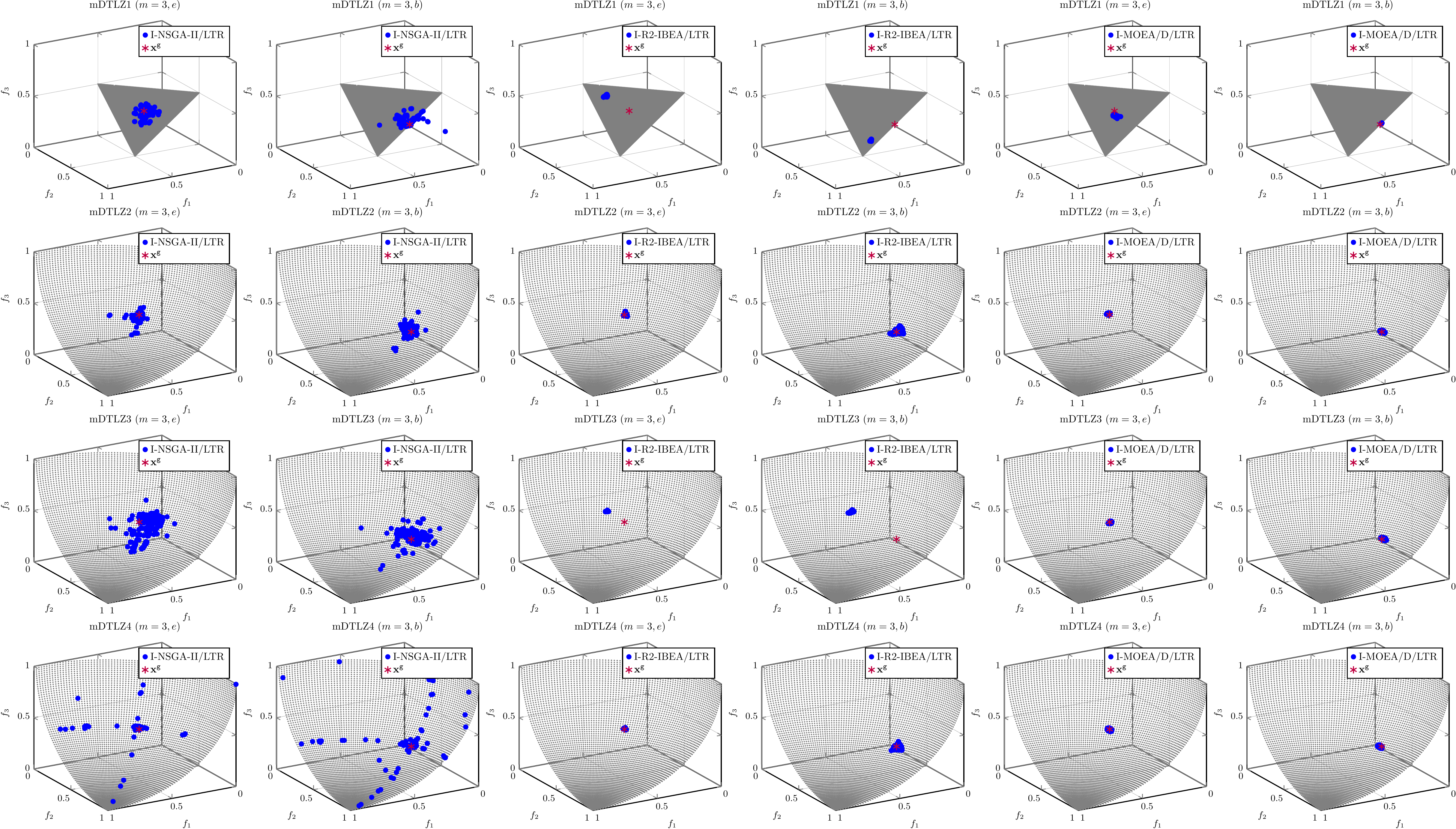}
    \caption{Population distribution of non-dominated solutions with the medium $\mathbb{E}(\mathcal{P})$ value obtained by \texttt{I-NSGA-II/LTR}, \texttt{I-R2-IBEA/LTR} and \texttt{I-MOEA/D/LTR} on mDTLZ1 to mDTLZ4, respectively, when $m=3$.}
    \label{fig:mdtlz}
\end{figure*}

Let us first look into the Wilcoxon signed-rank test results of $\mathbb{E}(\mathcal{P})$ shown in Tables 1 to 4 in the supplementary document. As shown in Table 1, we find that \texttt{I-NSGA-II/LTR} and \texttt{I-MOEA/D/LTR} are the most competitive algorithms on DTLZ1 to DTLZ6. In particular, all three proposed algorithm instances are more capable of finding the SOI w.r.t. an equal preference (i.e., $\mathbf{w}^\mathtt{e}$) than a biased preference (i.e., $\mathbf{w}^\mathtt{b}$). As the sample results shown in~\pref{fig:dtlz4_m3}, we can see that \texttt{I-NSGA-II/LTR} and \texttt{I-R2-IBEA/LTR} are struggling on DTLZ4, which is featured with an induced bias on certain objective functions. As the sample results shown in~\pref{fig:dtlz6_m8_m10}, we find that \texttt{I-NSGA-II/LTR} can hardly converge to the SOI when $m=8$ and $m=10$. DTLZ1$^{-1}$ to DTLZ4$^{-1}$ have an inverted PF shape as opposed to DTLZ1 to DTLZ4. From the results shown in Table 2, we find that \texttt{I-NSGA-II/LTR} has shown much better performance than the other two peer algorithms in most cases even for DTLZ4$^{-1}$ which is also featured with an induced bias as DTLZ4 (\pref{fig:minus_dtlz4_m10} gives the sample results on DTLZ4$^{-1}$ with $m=10$). mDTLZ1 to mDTLZ4 have the same PF shape as DTLZ1$^{-1}$ to DTLZ4$^{-1}$ but they are featured with hardly dominated boundary solutions. From the results shown in Table 3 and \pref{fig:mdtlz}, it is interesting to note that \texttt{I-MOEA/D/LTR} becomes a competitive peer algorithm in most cases. WFG3 is a challenging problem~\cite{IshibuchiMN16} whose PF is partially degenerated to a lower-dimensional manifold. As the results in Table 4 of the supplementary document and~\pref{fig:wfg3_m10}, we find that \texttt{I-NGSA-II/LTR} and \texttt{I-MOEA/D/LTR} have shown a better performance comparing to \texttt{I-R2-IBEA/LTR}.

\begin{figure*}[htbp]
    \centering
    \includegraphics[width=\linewidth]{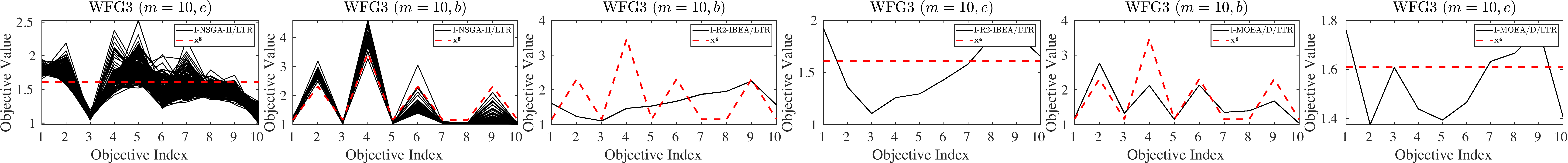}
    \caption{Population distribution of non-dominated solutions with the medium $\mathbb{E}(\mathcal{P})$ value obtained by \texttt{I-NSGA-II/LTR}, \texttt{I-R2-IBEA/LTR} and \texttt{I-MOEA/D/LTR} on WFG3 when $m=10$.}
    \label{fig:wfg3_m10}
\end{figure*}

\begin{figure}[htbp]
    \centering
    \includegraphics[width=.5\linewidth]{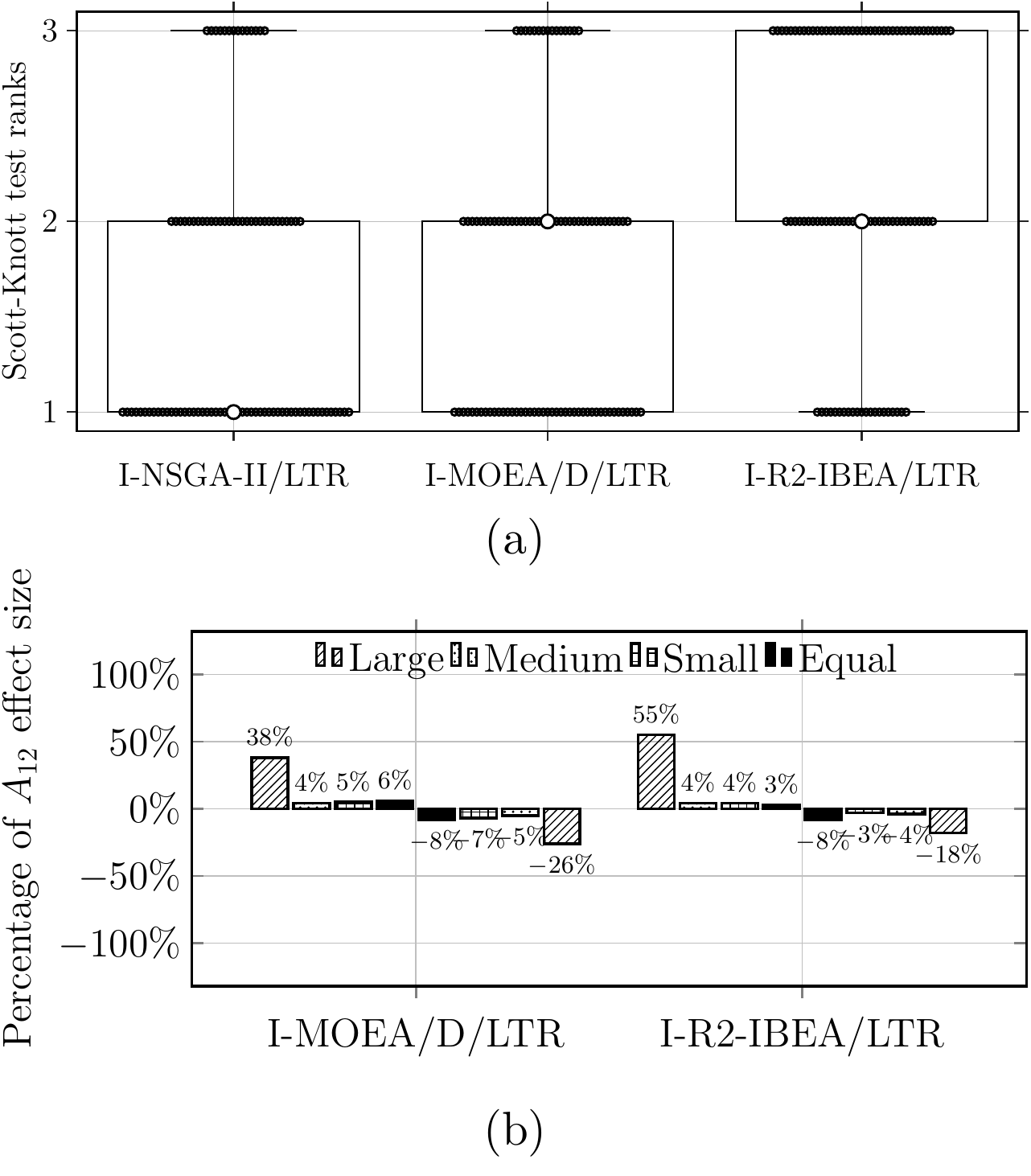}
    \caption{(a) Box plots for the Scott-knott test ranks achieved by each of the three algorithm instances of our proposed framework. The distribution of the obtained ranks are presented as the black $\newmoon$ symbol in the corresponding boxes. The median value is highlighted as a circle $\fullmoon$ in each box. (b) Percentage of the large, medium, small, and equal $A_{12}$ effect size of $\mathbb{E}(\mathcal{P})$ when comparing \texttt{I-NSGA-II/LTR} with \texttt{I-R2-IBEA/LTR} and \texttt{I-MOEA/D/LTR}. In particular, the negative part indicates the better $\mathbb{E}(\mathcal{P})$ achieved by \texttt{I-R2-IBEA/LTR} and \texttt{I-MOEA/D/LTR} against \texttt{I-NSGA-II/LTR}.}
    \label{fig:rq1_stats}
\end{figure}

In addition to the pairwise comparison conducted by the above Wilcoxon signed-rank test, we apply the Scott-knott test to facilitate a better ranking among three proposed algorithm instances. Due to the large number of test problem instances used in our experiments, it will be messy if we list all ranking results ($48\times2\times3=288$ in total) obtained by the Scott-knott test collectively. Instead, to have a better interpretation of the comparison among our proposed three algorithm instances based on LTR, we pull all the Scott-knott test results together and show their distribution and variance as the box plots in~\pref{fig:rq1_stats}(a). From this result, we find that \texttt{I-NSGA-II/LTR} is the best algorithm instance to approximate the SOI as it has been classified into the best group in most comparisons. In contrast, \texttt{I-R2-IBEA/LTR} is the worst algorithm instance. It is worth noting that both \texttt{I-R2-IBEA/LTR} and \texttt{I-MOEA/D/LTR} use the same preference elicitation method, i.e., a set of biased weight vectors representing the DM's preference information w.r.t. the ROI. In this case, the inferior results of \texttt{I-R2-IBEA/LTR} might be attributed to the relatively poor selection pressure provided by the $R2$ indicator.

As discussed above, \texttt{I-NSGA-II/LTR} stands out as the best algorithm instance of our proposed framework. To better understand the performance difference of \texttt{I-NSGA-II/LTR} w.r.t. \texttt{I-R2-IBEA/LTR} and \texttt{I-MOEA/D/LTR}, we investigate the comparison results of $A_{12}$ effect size between \texttt{I-NSGA-II/LTR} and the other two peer algorithms, respectively. From the bar charts shown in~\pref{fig:rq1_stats}(b), we find that the better results achieved by \texttt{I-NSGA-II/LTR} against \texttt{I-R2-IBEA/LTR} are more evident than \texttt{I-MOEA/D/LTR} given that more than half of the better results are classified to be statistically large. This also supports the observations from the Scott-knott test.

\begin{tcolorbox}[breakable, title after break=, height fixed for = none, colback = gray!40!white, boxrule = 0pt, sharpish corners, top = 0pt, bottom = 0pt, left = 2pt, right = 2pt]
    \underline{Answers to \textit{RQ}1}: We have the following takeaways from our experiments. 1) Among our proposed three algorithm instances, \texttt{I-NSGA-II/LTR} is the most competitive one for approximating the SOI in most cases. This is surprising at the first glance since the baseline NSGA-II is notorious for its poor scalability in problems with more than three objectives. 2) Although \texttt{I-R2-IBEA/LTR} and \texttt{I-MOEA/D/LTR} share the same preference elicitation method, the performance of \texttt{I-MOEA/D/LTR} is superior to that of \texttt{I-R2-IBEA/LTR}. This suggests that the Tchebycheff function can provide a stronger selection pressure than the $R2$ indicator. 3) The solutions approximated by \texttt{I-R2-IBEA/LTR} and \texttt{I-MOEA/D/LTR} are usually concentrated while those obtained by \texttt{I-NSGA-II/LTR} have a dispersed spread around the ROI. This difference is caused by their different preference elicitation methods.
\end{tcolorbox}

\subsection{Performance Comparison of the Three Proposed Algorithm Instances against the Other Peer Algorithms}
\label{sec:all_algorithms_comparisons}

In this subsection, we investigate the performance of our proposed three algorithm instances w.r.t. four peer algorithms. Let us again look into the Wilcoxon signed-rank test results of $\mathbb{E}(\mathcal{P})$ shown in Tables 5 to 8 in the supplementary document. Note that \texttt{NEMO-0} only has results for up to $8$ objectives. This can be partially attributed to the exponentially soaring complexity of the linear programming involved in \texttt{NEMO-0} that renders it extremely slow when the number of objectives becomes larger. From the comparison results, we can see that all our three proposed algorithm instances, even for the least competitive one \texttt{I-R2-IBEA/LTR}, have shown superior performance against the peer algorithms in most cases.

\begin{figure}[t!]
\centering
\includegraphics[width=.6\linewidth]{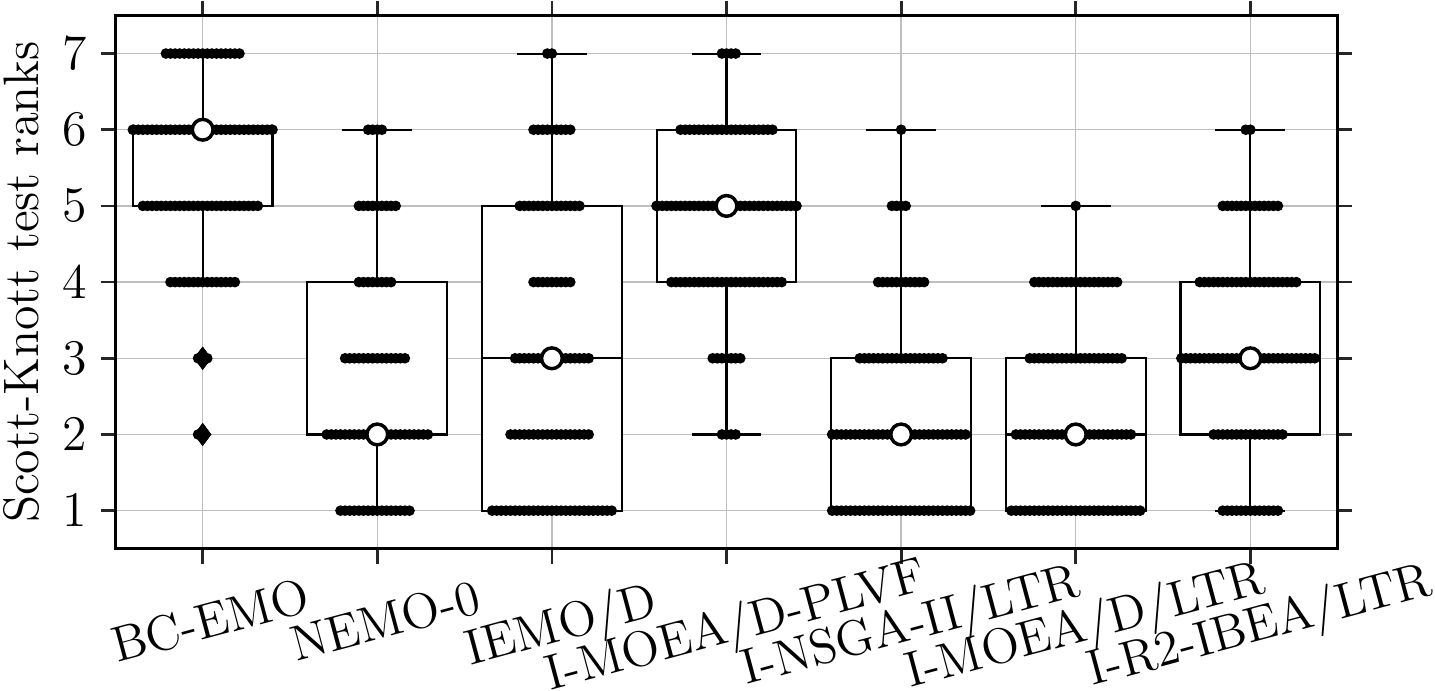}
\caption{Box plots for the Scott-knott test ranks achieved by each of the three algorithm instances of our proposed framework along with four peer algorithms. The distribution of the obtained ranks are presented as the black $\newmoon$ symbol in the corresponding boxes. The median value is highlighted as a circle $\fullmoon$ in each box.}
\label{fig:rq2_sk}
\end{figure}

Like in~\pref{sec:our_algorithms_comparisons}, we also apply the Scott-knott test to investigate the collected ranking relation ($48\times2\times7-11\times2=650$ in total) among our proposed algorithm instances and the peer algorithms. From the box plots shown in~\pref{fig:rq2_sk}, we can have a better view to see that all our three proposed algorithm instances are ranked as the most competitive algorithms while \texttt{I-NSGA-II/LTR} is still the best one. As for the selected peer algorithms, the performance of \texttt{NEMO-0} and \texttt{IEMO/D} are better than that of \texttt{I-MOEA/D-PLVF} and \texttt{BC-EMOA}.

\begin{figure}[t!]
\centering
\includegraphics[width=.6\linewidth]{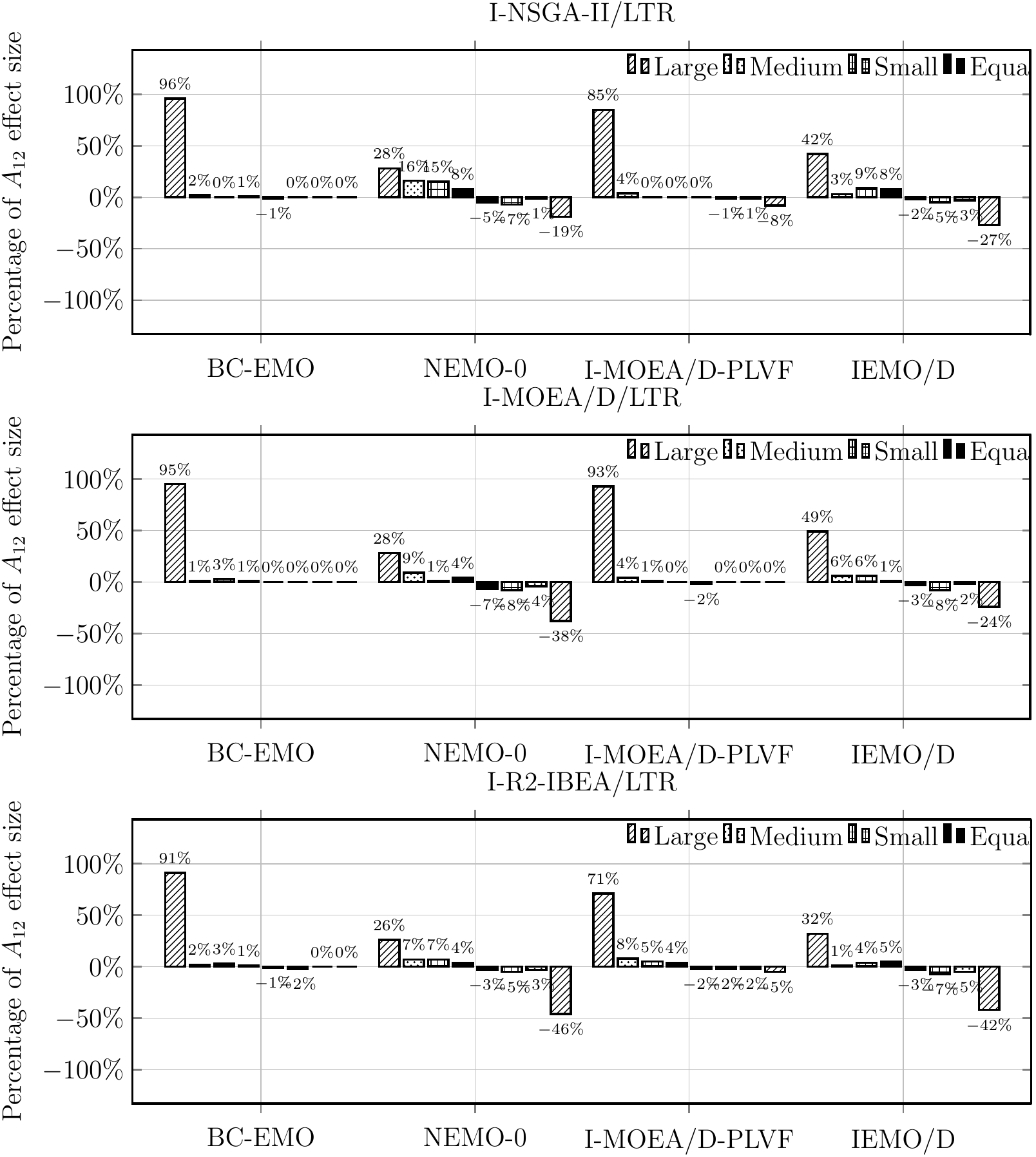}
\caption{Percentage of the large, medium and small $A_{12}$ effect size, respectively, when comparing each of our proposed three algorithm instances against the other four peer algorithms.}
\label{fig:rq2_a12}
\end{figure}

Last but not the least, we pick up each of our proposed three algorithm instances as a sentinel, respectively, and compare its performance difference with the other four peer algorithms by using the $A_{12}$ effect size. From the collected comparison results ($48\times2\times3\times3+37\times2\times3=1086$ in total) shown in~\pref{fig:rq2_a12}, we can see that all our three proposed algorithm instances have shown overwhelming advantages over \texttt{BC-EMOA} and \texttt{I-MOEA/D-PLVF} as the sum of the percentage of the effect size is always close to $90\%$. When comparing with \texttt{IEMO/D}, the advantages of our proposed algorithm instances have been narrowed down. This can be partially attributed to the competitive performance of \texttt{IEMO/D} when the number of objectives is small. As for \texttt{NEMO-0}, it has shown certain comparable performance with our proposed algorithm instances. In particular, it has achieved more superior results against \texttt{I-R2-IBEA/LTR} and \texttt{I-MOEA/D/LTR}. These are also largely because of the better performance achieved by \texttt{NEMO-0} when the number of objectives is small.

As expected, the performance of interactive EMO algorithms deteriorate with the increase of the number of objectives. However, it is interesting to find that NSGA-II, which is notorious for MOPs with more than three objectives, becomes surprisingly more resilient than MOEA/D as a baseline algorithm. This can be partially attributed to the increasing difficulties to: 1) learn a preference model due to the curse of dimensionality and short of human labeled data; and 2) specify appropriate weight vectors in a high-dimensional space.

\begin{tcolorbox}[breakable, title after break=, height fixed for = none, colback = gray!40!white, boxrule = 0pt, sharpish corners, top = 0pt, bottom = 0pt, left = 2pt, right = 2pt]
    \underline{Answers to \textit{RQ}2}: We have the following takeaways from our experiments. 1) All our three proposed algorithm instances have outperformed the other four peer algorithms in most comparisons. 2) Some of the selected peer algorithms, \texttt{IEMO/D} and \texttt{NEMO-0} in particular, have shown competitive performance when the number of objectives is small. 3) Comparing to MOEA/D, NSGA-II is a more resilient baseline algorithm to constitute an interactive EMO algorithm with the increase of the number of objectives.
\end{tcolorbox}

\subsection{Performance Comparison of the LTR Neural Network against the Other Peer Ranking Algorithms}
\label{sec:ranking_comparisons}

From the results discussed in Sections~\ref{sec:our_algorithms_comparisons} and~\ref{sec:all_algorithms_comparisons}, we confirm that \texttt{I-NGSA-II/LTR} is the most competitive algorithm for approximating the SOI. The core difference between \texttt{I-NGSA-II/LTR} and \texttt{BC-EMO} and \texttt{NEMO-0} (the selected peer algorithms in~\pref{sec:all_algorithms_comparisons} who also use NSGA-II as the baseline algorithm) is the model used to learn the ranking among candidate solutions. In particular, \texttt{BC-EMO} uses the ranking SVM while \texttt{NEMO-0} uses the ordinal regression. To address RQ3, we first create a set of synthetic data $\mathcal{S}=\{\mathbf{t}^i\}_{i=1}^N$. In particular, $\mathbf{t}=(t_1,\cdots,t_m)^T$ where $t_j$ is uniformly sampled from $[0,1]$, $j\in\{1,\cdots,m\}$ and we consider $m\in\{2,3,5,8,10\}$ respectively in our experiment. During the experiment, we randomly pick up $50$ sample pairs from $\mathcal{S}$ to constitute the training dataset for each LTR algorithm. To have a quantitative comparison, we apply the normalized discounted cumulative gain (NDCG) metric~\cite{JarvelinK17}, widely used in the information retrieval domain, to evaluate the LTR performance.

From the comparison results shown in~\pref{tab:rq3_ranking_comparisons}, it is clear to see that the ranking SVM is the worst ranking algorithm. This explains the inferior performance of \texttt{BC-EMO} as discussed in~\pref{sec:all_algorithms_comparisons}. In contrast, the performance of ordinal regression and LTR neural network is comparable. This observation also confirms the competitive results of \texttt{NEMO-0} for approximating the SOI as shown in~\pref{fig:rq2_a12}. However, we argue that the ordinal regression in \texttt{NEMO-0} is not flexible enough to represent the DM's preference information. In Appendix A of this paper\footnote{The appendix of this paper can be found in~\url{https://tinyurl.com/52856m6b}.}, we present a counterexample that shows the inability of the ordinal regression to learn an appropriate additive value function even when the DM provides accurate and consistent preference information.

\begin{table*}[t!]
    \centering
    \caption{Comparison results on NDCG@20 metric (median and IQR) for LTR neural network and the peer ranking algorithms}
    \label{tab:rq3_ranking_comparisons}
    \resizebox{.8\textwidth}{!}{ 
    \begin{tabular}{c|c|c|c|c|c}
        \hline
        Ranking algorithm & $m=2$   & $m=3$   & $m=5$   & $m=8$   & $m=10$ \\ \hline
        Ranking SVM & 0.1954(1.74E-2)$^\dag$  & 0.6425(1.21E-1)$^\dag$ & 0.6801(6.51E-2)$^\dag$ & 0.8421(3.53E-2)$^\dag$ & 0.8360(3.67E-2)$^\dag$ \\ \hline
        Ordinal regression & 0.9840(1.99E-2)$^\dag$  & 0.9814(1.19E-2)$^\dag$  & 0.9452(4.07E-2)$^\dag$ & 0.9200(4.14E-2)$^\dag$ & 0.9290(2.88E-2)$^\dag$ \\ \hline
        LTR neural network & \cellcolor[rgb]{ .749,  .749,  .749}\textbf{0.9984(3.11E-3) } & \cellcolor[rgb]{ .749,  .749,  .749}\textbf{0.9827(4.26E-2) } & \cellcolor[rgb]{ .749,  .749,  .749}\textbf{0.9698(1.13E-2)} & \cellcolor[rgb]{ .749,  .749,  .749}\textbf{0.9409(1.93E-2)} & \cellcolor[rgb]{ .749,  .749,  .749}\textbf{0.9340(4.02E-2) } \\ \hline
    \end{tabular}
    }
    \begin{tablenotes}
    \item[1] $^{\dag}$ denotes the performance of LTR neural network is significantly better than the other peers according to the Wilcoxon's rank sum test at a 0.05 significance level; $^{\ddag}$ denotes the corresponding algorithm significantly outperforms the LTR neural network. NDCG@20 indicates that the DM is only interested in the correctness of the top-$20$ ranking results.
    \end{tablenotes} 
\end{table*}

\begin{tcolorbox}[breakable, title after break=, height fixed for = none, colback = gray!40!white, boxrule = 0pt, sharpish corners, top = 0pt, bottom = 0pt, left = 2pt, right = 2pt]
    \underline{Answers to \textit{RQ}3}: We have the following takeaways from this experiment. 1) The performance of a LTR algorithm can have impacts to the effectiveness of the interactive EMO algorithm. 2) Ranking SVM is less capable than the LTR neural network and the ordinal regression. However, the latter is not robust to learn an appropriate additive value function, which finally renders its ineffectiveness.
\end{tcolorbox}

\subsection{Sensitivity Study of Parameters}
\label{sec:parameter_sensitivity}

There are three parameters associated with our proposed interactive EMO framework based on a LTR neural network, i.e., $\mu$, $\tau$ and $\eta$ as introduced in~\pref{sec:parameters}. In particular, $\mu$ and $\tau$ are only related to \texttt{I-NSGA-II/LTR} while \texttt{I-MOEA/D/LTR} and \texttt{I-R2-IBEA/LTR} involve all three parameters. In this subsection, we plan to empirically investigate the sensitivity of the performance of our three algorithm instances w.r.t. these parameters. To this end, we consider different settings of $\tau=\{5,10,20\}$, $\mu=\{5,10,20\}$ and $\eta=\{0.1,0.2,0.4\}$ while the other parameters are kept the same as introduced in~\pref{sec:parameters} and the experiments are conducted on the benchmark problems as introduced in~\pref{sec:benchmark}.

\begin{figure*}[t!]
	\centering
	\includegraphics[width=\linewidth]{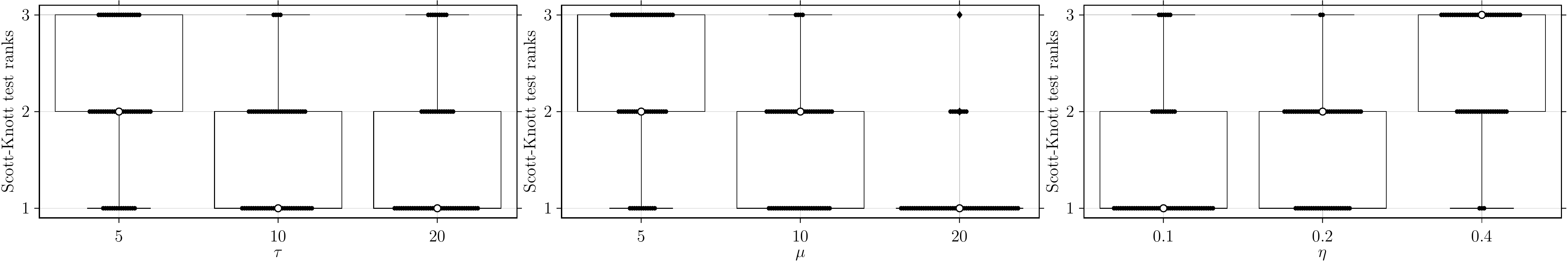}
	\caption{Box plots of Scott-knott test ranks achieved by different settings of $\tau$, $\mu$ and $\eta$ (the smaller rank is, the better performance is achieved). The distribution of the obtained ranks are presented as the black $\newmoon$ symbol in the corresponding boxes. The median value is highlighted as a circle $\fullmoon$ in each box.}
	\label{fig:rq4_box_plots}
\end{figure*}

\subsubsection{Effect of $\tau$}
It controls the number of generations between two consecutive consultation sessions. Specifically, a small $\tau$ means that we need to frequently consult the DM about pairwise comparisons. By doing so, we can in principle expect a more accurate preference model due to the increased amount of labeled data. However, from the box plots shown in~\pref{fig:rq4_box_plots}, it is surprising to see that the overall $\mathbb{E}(P)$ values degenerate with the increase of $\tau$. This can be attributed to the too frequent adjustment of the preference model at the beginning of the search process. It is detrimental to mislead the EMO algorithm due to the less accurate preference information.

\subsubsection{Effect of $\mu$}
As introduced in~\pref{sec:consultation}, $\mu$ determines the number of labeled data (pairwise comparisons made by the DM) that can be used to train the LTR neural network. It makes sense that the more data the DM can provide, the more accurate LTR model you can expect. This assumption is supported by the box plots shown in~\pref{fig:rq4_box_plots}, \texttt{I-MOEA/D/LTR} can have a better performance when using a large $\mu$. However, presenting the DM too many solutions for pairwise comparisons will inevitably increase her workload, thus leading to her fatigue. On the other hand, the model accuracy will be impaired if the data are significantly insufficient.

\subsubsection{Effect of $\eta$}
As introduced in~\pref{sec:elicition}, $\eta$ controls the convergence rate of the weight vectors towards the promising ones identified by the LTR neural network learned from the consultation module. A large $\eta$ leads a fast convergence but may have a risk of premature convergence towards an undesired region. On the contrary, a small $\eta$ may slow down the convergence towards the SOI within the limited number of FEs. It can be observed from the box plots in~\pref{fig:rq4_box_plots} that a smaller $\eta$ leads to better results in most cases.

\begin{tcolorbox}[breakable, title after break=, height fixed for = none, colback = gray!40!white, boxrule = 0pt, sharpish corners, top = 0pt, bottom = 0pt, left = 2pt, right = 2pt]
    \underline{Answers to \textit{RQ}4}: We have the following takeaways from our experiments. 1) It is not recommended to consult the DM too frequently as this not only increases the risk of making the DM fatigue but also introduces extra noises to the search process. 2) It is anticipated to improve the accuracy of the LTR model by involving more labeled data from the DM. However, this brings extra queries that inevitably increase the DM's workloads. 3) A small step size $\eta$ can be beneficial to fine tune the search direction towards the SOI. However, this may also lead to a slow convergence towards the SOI.
\end{tcolorbox}

\subsection{Impact of Inconsistency in the Elicited Preference}
\label{sec:inconsistency}

\begin{figure}[t!]
    \centering
    \includegraphics[width=.5\linewidth]{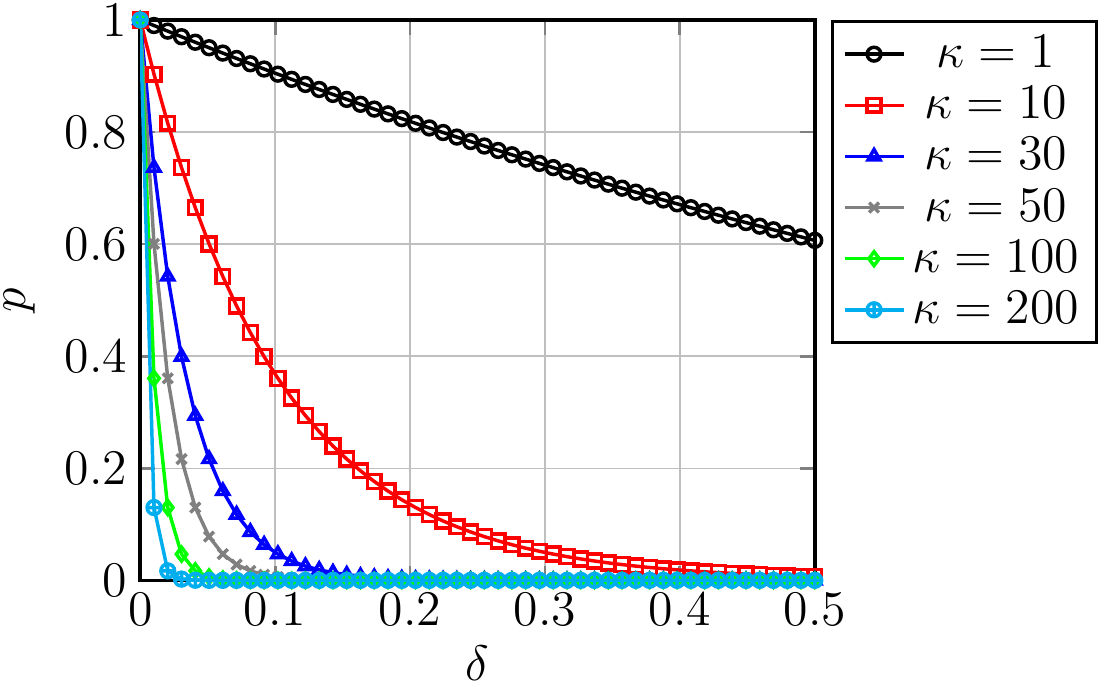}
    \caption{An illustrative example of the impact of the noise level, i.e., different settings $\kappa$, on $\mathbb{P}(\mathbf{x}^i,\mathbf{x}^j)$.}
    \label{fig:rq5_func_fig}
\end{figure}

In the previous experiments, the DM's preference information is assumed to be deterministic. However, it is not uncommon that practical decision-making and preference elicitation can be largely inconsistent. In other words, there exist certain level of noises to which the pairwise comparison results can be conflicting w.r.t. the ground truth. In this subsection, we plan to investigate the impact brought by the inconsistencies in the preference elicitation. To this end, our basic idea is to aggregate a random error into the pairwise comparison. Specifically, given a pair of selected solutions $\langle\mathbf{x}^i,\mathbf{x}^j\rangle$, we define the probability of flipping the comparison result as:
\begin{equation}
    \mathbb{P}(\mathbf{x}^i,\mathbf{x}^j)=\exp(-\kappa\cdot\delta),
\end{equation}
where $\kappa$ determines the DM's ability to correctly express her preference information and $\delta=\left|\psi(\mathbf{x}^j)-\psi(\mathbf{x}^i)\right|$ measures the \lq similarity\rq\ of a given pair of solutions. From the illustrative example shown in~\pref{fig:rq5_func_fig}, we can infer that a smaller $\kappa$ leads to a larger chance of eliciting a wrong preference information, i.e., the DM picks up an inferior solution as the winner from the pairwise comparison. Moreover, a larger $\delta$ indicates a more obvious difference between the given solution pair. In principle, a rationale DM is assumed to be less likely to make a wrong decision if the candidates are obviously different, and vice versa. In our experiment, we investigate different settings of $\kappa\in\{1,10,30,50,100,200\}$. For proof-of-concept purposes, DTLZ2 and mDTLZ2 are chosen as the benchmark test problems and the parameters associated with our algorithm instances are kept the same as introduced in~\pref{sec:parameters}.

\begin{figure*}[t!]
    \centering
    \includegraphics[width=\linewidth]{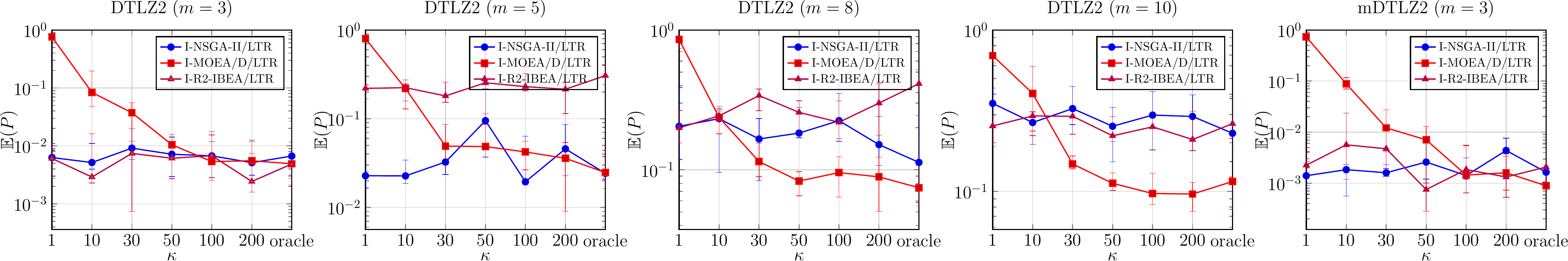}
    \caption{Comparison results of the variance of the approximation errors with difference $\kappa$ settings.}
    \label{fig:rq5_variance}
\end{figure*}

From the comparison results shown in~\pref{fig:rq5_variance}, we can see the performance of \texttt{I-MOEA/D/LTR} is influenced by the induced noise in the preference elicitation. Its approximation error w.r.t. the SOI is large when involving a large noise in the preference elicitation (i.e., having a small $\kappa$); while the approximation accuracy is very close to the noiseless case (denoted as the \lq oracle\rq\ in~\pref{fig:rq5_variance}) with the increase of $\kappa$ (i.e., the noise in the preference elicitation becomes trivial). These observations indicate that \texttt{I-MOEA/D/LTR} has certain level of robustness w.r.t. mild inconsistencies during the preference elicitation. On the other hand, it is interesting to note that the induced noise in the preference elicitation does not pose significant impacts to the performance of \texttt{I-NSGA-II/LTR} and \texttt{I-R2-IBEA/LTR}. In particular, since \texttt{I-R2-IBEA/LTR} shares the same preference elicitation method with \texttt{I-MOEA/D/LTR}, we infer the differences of robustness w.r.t. the noise are derived from the environmental selection in MOEA/D and R2-IBEA. This can be explained as the evolutionary search process is less dependent on the preference information for both \texttt{I-NSGA-II/LTR} and \texttt{I-R2-IBEA/LTR}. Therefore, both of them are able to drive the population towards the PF even with a large noise.

\begin{tcolorbox}[breakable, title after break=, height fixed for = none, colback = gray!40!white, boxrule = 0pt, sharpish corners, top = 0pt, bottom = 0pt, left = 2pt, right = 2pt]
    \underline{Answers to \textit{RQ5}}: There are two interesting takeaways from the experiments in this subsection. 1) The performance of \texttt{I-NSGA-II/LTR} and \texttt{I-R2-IBEA/LTR} are resilient to the induced noise in the preference elicitation. 2) \texttt{I-MOEA/D/LTR} can be impaired by involving a large noise in the preference elicitation but it is still robust to some mild inconsistencies.
\end{tcolorbox}


\section{Conclusion}
\label{sec:conclusion}

This paper developed a general framework to design interactive EMO algorithms that progressively learn the DM’s preference information from her feedback and adapt the learned preference to guide the population towards the SOI.
It consists of three modules, i.e., \texttt{consultation}, \texttt{preference elicitation} and \texttt{optimization}. As an interface between the DM and the EMO algorithm, the \texttt{consultation} module collects the implicit preference information (in the form of pairwise comparisons), based on which it learns a latent preference model. Once the DM's latent preference information is learned, the \texttt{preference elicitation} module translates it into a tailored form that can be used in any prevalent EMO algorithms according to their environmental selection mechanisms. Extensive experiments on three- to ten-objective test problems fully demonstrated the effectiveness of our proposed framework for helping three iconic EMO algorithms for finding the DM's preferred solution(s).

As discussed in~\cite{LiLDMY20}, the synergy of ideas between EMO and MCDM is an exciting direction to push the boundary of multi-objective optimization and decision-making. This paper can be extended from at least three aspects. First, since the physical queries can be labor-intensive and error-prone, it is interesting to investigate active learning mechanisms~\cite{RenXCHLGCW22,ChenLTL22,Chen021a,FanLT20} to pick up the most informative samples in a strategic manner. In view of the black box nature of real-world optimization problems, the explainability have rarely been explored in the literature. It is worthwhile to investigate the interpretability of the obtained SOI along with the trade-off among conflicting objectives~\cite{LiK14,NieGL20,CovertLL21,Chen021,LiNGY21}. This can facilitate the understanding of the DM's latent preference information and further advance a better informed MCDM. Last but not the least, advanced techniques developed in the human-computer interaction domain can be leveraged to realize a human-machine symbiosis in future~\cite{MullerS15,GaoNL19}.

\section*{Acknowledgment}
This work was supported by UKRI Future Leaders Fellowship (MR/S017062/1), EPSRC (2404317), NSFC (62076056), Royal Society (IES/R2/212077) and Amazon Research Award.

\bibliographystyle{IEEEtran}
\bibliography{IEEEabrv,ranknet}

\end{document}